\pdfoutput=1

\documentclass[11pt]{article}

\usepackage[final]{acl}

\usepackage{booktabs}
\usepackage{todonotes}
\usepackage{times}
\usepackage{latexsym}
\usepackage{multirow}
\usepackage{xcolor}
\usepackage{mathtools,mathrsfs}
\usepackage{amsmath, amssymb}
\usepackage{graphicx}
\usepackage{caption}
\usepackage{subcaption}
\usepackage{enumitem}
\usepackage{appendix}
\usepackage{amsmath}
\usepackage{paralist}
\usepackage{bm}
\usepackage[table]{xcolor}

\usepackage[T1]{fontenc}
\usepackage{listings}

\usepackage[utf8]{inputenc}

\usepackage{microtype}

\usepackage{inconsolata}
\newcommand{\blue}{\textcolor{black}}
\definecolor{Daniil}{RGB}{123, 45, 67}

\newcommand{\us}[1]{\textcolor{black}{#1}}

\title{A Multi-View Media Profiling Suite: Resources, Evaluation, and Analysis}

\author{
Muhammad Arslan Manzoor$^{\alpha,\beta}$\thanks{These authors contributed equally to this work.} \quad
Dilshod Azizov$^{\alpha}$\footnotemark[1] \quad
Daniil Orel$^{\alpha}$\footnotemark[1] \quad \\
\textbf{Umer Siddique}$^{\gamma}$ \quad
\textbf{Zain Muhammad Mujahid$^{\delta}$ \quad
Yufang Hou$^{\beta}$ \quad
Preslav Nakov$^{\alpha}$} \\
\\
$^{\alpha}$MBZUAI, UAE \quad
$^{\beta}$Interdisciplinary Transformation University, Austria \\
$^{\gamma}$University of Texas at San Antonio, USA \quad
$^{\delta}$University of Copenhagen, Denmark \\
\texttt{\{muhammad.arslan, preslav.nakov\}@mbzuai.ac.ae}
}

\begin{document}

\maketitle

\begin{abstract}

News outlets shape public opinion on a scale that makes automated detection of political bias and factuality essential. Yet, the field still lacks unified resources, comprehensive evaluations in diverse approaches, and systematic analyzes of the representations and fusion strategies that matter the most, especially under label sparsity and dataset diversity. In addition, there is little empirical work that reports broad observation-driven findings about what consistently works, what fails, and why. We address these gaps with four contributions: \emph{(i)} \textbf{MBFC-2025}, a large-scale label set that covers $\sim$2,600 outlets from Media Bias/Fact Check (MBFC); \emph{(ii)} multi-view representations for ACL-2020~\cite{EMNLP2022:GREENER} ($\sim$900 outlets) and MBFC-2025, spanning Alexa graphs, hyperlink graphs, LLM-derived graphs, articles, and Wikipedia descriptions; \emph{(iii)} systematic evaluation and analysis of embedding views and fusion strategies, including an RL-based fusion variant; and \emph{(iv)} extensive experiments that achieve state-of-the-art results on ACL-2020 and establish strong benchmarks on MBFC-2025.

\end{abstract}

\section{Introduction}

Profiling of news outlets has become increasingly important~\cite{baly-etal-2020-written,EMNLP2022:GREENER,mehta2023interactive,mehta2022tackling} because claim- or article-level verification is costly and difficult to scale. Outlet-level profiling offers a practical alternative: by characterizing sources through their historical \emph{political bias} and \emph{factuality}, it enables scalable, real-time monitoring. A widely used source of supervision is Media Bias/Fact Check (MBFC), which provides expert outlet labels and has served as the ground truth in many studies~\cite{mehta2023interactive,EMNLP2022:GREENER,baly2018predicting,baly2019multi,hounsel2020identifying,nakov2021survey}. However, effective profiling depends on reliable outlet representations.

Most existing methods rely on textual content~\cite{baly-etal-2020-detect,baly2019multi}, which can be noisy (frame, boilerplate), difficult to crawl, and expensive to collect on scale. Thus, recent work explores alternative signals, including infrastructure and website metadata~\cite{hounsel2020identifying}, social context such as followers~\cite{baly-etal-2020-written}, and graph-based representations that connect outlets through audiences or hyperlinks~\cite{EMNLP2022:GREENER,manzoor2025mgm}. While promising, graph-based approaches can suffer from structural sparsity (\emph{e.g.,} disconnected components) that limits representation learning~\cite{longa2024explaining,zhang2024linear}, and prior evaluations are often tied to a single graph view and benchmark, leaving open which signals and fusion strategies are most reliable under missing or noisy modalities.

To fill these gaps, we build a \emph{multi-view} profiling suite that combines three graph views and two textual views. The graph views include: \emph{(i) Alexa graphs} capturing audience overlap, \emph{(ii) hyperlink graphs} reflecting inter-outlet links, and \emph{(iii) LLM-derived graphs} encoding semantic similarity between outlets. The textual views include: \emph{(iv) outlet articles} capturing tone and framing, and \emph{(v) Wikipedia descriptions} providing context. We evaluate these views in the ACL-2020~\cite{EMNLP2022:GREENER} label set, which covers fewer than 900 outlets with 3-point political bias and factuality scales, and where many outlets lack article- or Wikipedia-based representations due to missing content and crawling constraints. To improve coverage and granularity, we introduce \textbf{MBFC-2025}, a source-level benchmark of $\sim$2{,}6k outlets with 5-point scale labels for both tasks. We construct the same multi-view representations for MBFC-2025 and evaluate embedding and fusion strategies to identify effective combinations. We also include a reinforcement learning (RL)-based fusion variant, formulated as a contextual bandit, to explore dynamic weighting under noisy and incomplete views.

We train graph neural networks (GNNs) over the media graphs and fine-tune pretrained language models (PLMs) on articles and Wikipedia entries. Our empirical study provides strong baselines, clarifies which representations contribute the most, and yields state-of-the-art results on ACL-2020 while establishing new benchmarks on MBFC-2025.

Our contributions are as follows:
\begin{compactitem}
    \item \textbf{Benchmark.} We introduce \textbf{MBFC-2025}, a large-scale source-level benchmark annotated by MBFC with $\sim$2{,}600 outlets labeled for political bias and factuality on a 5-point scale.
    
    \item \textbf{Resources.} We compile five-view representations for ACL-2020 and MBFC-2025: three graphs and two text representations.

    \item \textbf{Evaluation and Analysis.} We systematically evaluate embedding views and fusion strategies (including an RL-based variant), analyze what works and why, achieve state-of-the-art results on ACL-2020, and establish strong benchmarks on MBFC-2025\footnote{Our code can be accessed at the following github repository: \href{https://github.com/marslanm/multi-graph-perspective}{https://github.com/marslanm/multi-graph-perspective}.}.
\end{compactitem}


\section{Related Work}
\label{sec_related_work}

\textbf{Political bias:} Political bias reflects the ideological orientation of text that influences opinions and voting behavior \cite{druckman2005impact, boyle2007ideology, prior2013media}. Early work focused on textual analysis \cite{afroz2012detecting, perez2017automatic, conroy2015automatic, DBLP:conf/clef/MartinoAHNAN23, DBLP:conf/ecir/BarronCedenoACMEGHRSNCAN23, DBLP:conf/clef/BarronCedenoAGMNEACCHHKLRSZ23, DBLP:conf/clef/AzizovNL23}, while later studies incorporated contextual signals, including multimedia, infrastructure, and social context \cite{baly-etal-2020-written, hounsel2020identifying, castelo2019topic, fairbanks2018credibility}. Recent work uses PLMs and multimodal data, including BERT-based models \cite{guo2022measuring}, text-image methods \cite{qiu-etal-2022-late}, and ideology-driven pretraining \cite{liu-etal-2022-politics}. More recently, prompt-based methods \cite{maab2024media, mujahid2025profiling}, adaptable representations \cite{lin2024indivec}, and cross-lingual techniques \cite{DBLP:conf/emnlp/AzizovMANL24} have been explored.

The reliability of the media is commonly assessed by aligning the predictions of the model with the expert ratings \cite{yang2023large}. Earlier approaches relied on positions towards claims \cite{mukherjee2015leveraging, Popat:2017:TLE:3041021.3055133}, while later work used gold labels and PLMs for outlet profiling \cite{baly-etal-2020-written}.

\citet{mehta2023interactive} combine graph models, PLMs, and human input for fake news detection. RL has been explored for reliability prediction \cite{burdisso2024reliability}, but primarily for scoring rather than representation fusion.

\noindent\textbf{GNNs:} GNNs are used for representation learning in media analysis \cite{mehta2022tackling, manzoor2025mgm}. \citet{EMNLP2022:GREENER} construct homophilic graphs showing that similar outlets attract audiences. \citet{mehta2023interactive} integrate graph models, LLMs, and human input without requiring labels. Other work incorporates multihop reasoning \cite{zhang2022kcd} and event graphs \cite{lei2024sentence} to improve bias detection.

\noindent\textbf{Fusion:} Media graphs suffer from sparsity and few labels, hindering GNNs from capturing dependencies \cite{cui2020adaptive, sun2020multi, DBLP:conf/icml/BodnarF0OMLB21}. Previous work addresses this through multi-view learning, constructing graphs, and combining embeddings via strategies such as concatenation, averaging, or attention \cite{ding2022data,ma2020multi,yuan2021semi}. In contrast, we propose an RL-based fusion framework that dynamically learns outlet-specific weights over heterogeneous graph and textual embeddings. To our knowledge, this is the first application of RL for multi-view fusion in bias and factuality detection.

\section{Data Construction Pipeline}
We use MBFC as the label source for both datasets (details are in the Appendix~\ref{sec:appendix}). ACL-2020~\citep{baly-etal-2020-written} includes 859 media sources with 3-point labels for political bias and factuality (\emph{left/center/right}, \emph{high/mixed/low}), while MBFC-2025 uses a 5-point scale, adding \emph{left-center}, \emph{right-center}, \emph{very low}, and \emph{very high}. Next, we describe the representations and dataset statistics.

\subsection{Graphs Construction}
Generating diverse graphs and textual embeddings is key to address feature scarcity and enable data-driven models to better understand the underlying relationships and patterns. 

\begin{figure*}[t!]
    \centering
    \includegraphics[width=1.0\linewidth]{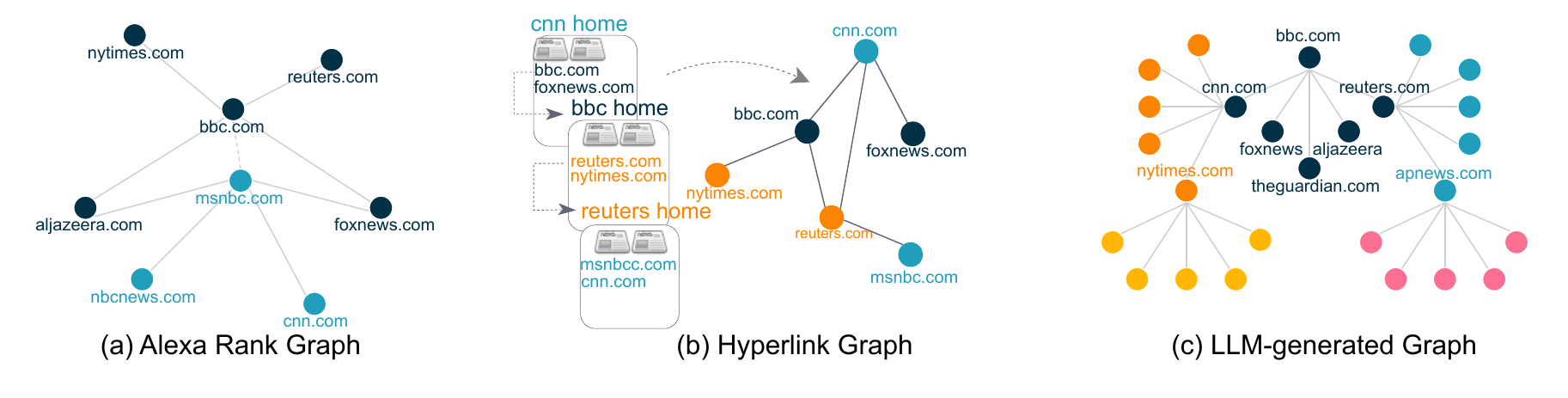}
    \caption{Illustration of generated graphs using (a) Alexa Rank tool (b) Hyperlink graph, and (c) LLM.}
    \label{graphs}
\end{figure*}

In addition to the Alexa graph, we construct two new graphs: the Hyperlink graph and the LLM-graph, illustrated in Figure~\ref{graphs}. They capture both inherent and implicit relationships between nodes. These relationships enable the generation of rich embeddings that ultimately complement other representations, \emph{e.g.,} text-based embeddings, by addressing knowledge gaps.

\begin{figure}[!t]
    \centering
    \resizebox{0.9\linewidth}{!}{
        \includegraphics{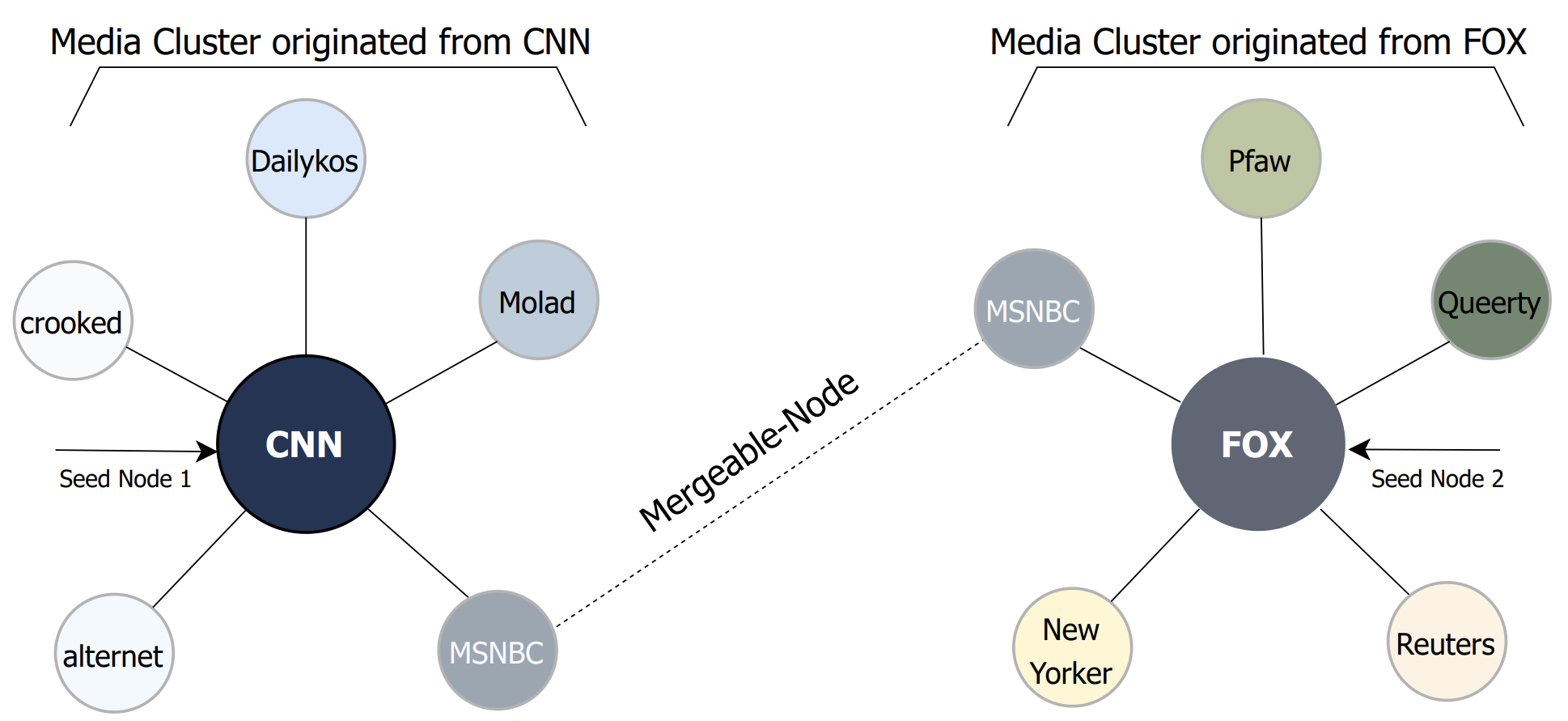}
    }
    \caption{CNN and FOX are seed nodes in the labeled media set (ACL-2020 \& MBFC-2025), while MSNBC serves as a mergeable node, forming a fundamental building block for graph construction.}
    \label{merge_fig}
\end{figure}

\textbf{Alexa Graph.} \citet{EMNLP2022:GREENER} used the Alexa Rank tool\footnote{\url{http://www.alexa.com/siteinfo}} to query the addresses of the news media, generating a list of 4-5 similar sites based on the overlap of the audience. We used these similar sites as nodes in a graph, with audience overlap defining the edges. Starting with websites from the ACL-2020 dataset as seed nodes, they iteratively expanded the graph by querying for each website and adding new nodes and edges. This iterative process was repeated five times, resulting in five levels of graphs, each progressively denser and more comprehensive.

Analysis of the constructed graphs in the Appendix \ref{stat_of_alexa} revealed several disconnected components, indicating isolated networks of nodes. As the graph levels increased, the number of disconnected components decreased, likely due to the growing number of nodes facilitating component merging. 

Graph level 3, the highest publicly accessible level for both factuality and political bias tasks, was chosen to produce embeddings with GNNs. The Alexa Rank tool also provided node representations, treated as node attributes during GNN training. These representations, including site rank, total linked sites, bounce rate, and daily user time \cite{EMNLP2022:GREENER}.

\textbf{Hyperlink (On-site) Graph.} To collect hyperlinks, we crawled with the newspaper3k\footnote{\url{https://github.com/codelucas/newspaper}} parser and extracted external links. We scraped the main page and articles, collecting up to 50 articles per site.
The nodes represent unique websites in the network, while the edges are counted twice for each pair of websites, since every link from website A to B creates two edges: A→B and B→A. To address graph sparsity, we employed a layered parsing strategy \us{that iteratively expands from dataset websites (level 0) to linked sites at subsequent levels, where level 0 includes only websites from the dataset, and each additional level incorporates websites referenced by those in the previous level.}

\textbf{LLM-Graph}
Techniques such as Chain-of-Thought (CoT) prompting \cite{wei2022chain}, Least-to-Most prompting \cite{zhou2022least}, and Tree-of-Thought (ToT) \cite{yao2024tree}, use LLM reasoning to tackle complex tasks. However, to leverage the contextual knowledge of LLMs for this task, we prompt the model to retrieve five websites similar to a given media outlet \(x\). The intuition behind this approach is two-fold: First, rather than directly asking the LLM to profile a media source, we leverage its contextual understanding to generate smaller clusters, seeded from ACL-2020 and MBFC-2025 datasets, which contain media labeled for factuality and political bias.

These clusters serve as the foundation for constructing media graphs, where we identify and merge the same nodes as shown in Figure \ref{merge_fig}. Second, this approach maintains consistency with other graph construction methods, where the number of similar nodes retrieved \(n\) is limited to five or fewer. To model the relationship between media outlets, we use GNNs, which generate rich media representations that are aware of the broader media landscape, including connections and distances between different media.

To implement this, we used the OpenAI \texttt{Python} package to query the \texttt{API} endpoint of the \textit{gpt-3.5-turbo-0125} model (GPT-3.5), released on January 25, 2024 \cite{NEURIPS2022_b1efde53, brown2020language}. The specific prompt used is as follows:

\begin{lstlisting}[language=Python,frame=lines,framesep=1mm,basicstyle=\ttfamily\footnotesize\linespread{1.1}\selectfont,mathescape=true,breaklines=true,showstringspaces=false,aboveskip=1em,belowskip=1em]
prompt = "Based on similarity, give me 5 similar websites to, {domain}. Return only the websites URL, strictly without any explanation, don't add numbers in start. Each website should be wrapped under the tag <s>"
\end{lstlisting}
In this prompt, the \textbf{\{domain\}} placeholder is replaced with the domain of interest, such as \emph{foxnews.com}. A sample response derived from GPT-3.5 for \emph{bbc.com} is shown below. 

\begin{lstlisting}[language={},frame=lines,framesep=1mm,basicstyle=\ttfamily\footnotesize\linespread{1.5}\selectfont,mathescape=true,breaklines=true,showstringspaces=false,aboveskip=2em,belowskip=2em]
<s>https://www.cnn.com/</s>
<s>https://www.theguardian.com/</s>
<s>https://www.aljazeera.com/</s>
<s>https://www.nytimes.com/</s>
<s>https://www.reuters.com/</s>
\end{lstlisting}

\begin{table}[!t]
\centering
\resizebox{0.49\textwidth}{!}{%
\begin{tabular}{l|ccccc|c}
\toprule
     \textbf{Political Bias}  & \textbf{Left} & \textbf{Left-center} & \textbf{Center} & \textbf{Right-center} & \textbf{Right} & \textbf{Total} \\
\midrule
ACL-2020 & 152 & - & 178 & - & 145 & 475 \\
MBFC-2025 & 255 & 427 & 705 & 759 & 404 & 2550 \\
\toprule
   \textbf{Factuality} & \textbf{Very High} & \textbf{High} & \textbf{Mixed} & \textbf{Low} & \textbf{Very Low} & \textbf{} \\
\toprule
ACL-2020  & - & 295 & 119 & 61 &  - & 475   \\
MBFC-2025 & 30 & 1347 & 970 & 147 & 56 & 2550    \\

\bottomrule
\end{tabular}
}
\vspace{-0.5em}
\caption{Label distribution statistics for political bias and factuality in the ACL-2020 (3-point scale) and MBFC-2025 (5-point scale) datasets.}
\label{label distribution}
\end{table}

We convert responses into JSONL for analysis and storage, then build a deduplicated level-0 graph and recursively expand it to level 3. The resulting level-3 graph contains up to 18,508 nodes for ACL-2020 and 80,385 for MBFC-2025, spanning 475 and 2{,}550 labels, respectively, as shown in Table~\ref{label distribution} and Table~\ref{tab:combined-stats}. Before scaling, we manually validated a subset of MBFC outputs. Since MBFC media are the LLM inputs, we profile the input media rather than generated URLs, preserving a connected graph for structural embeddings and robustness to sparsity.

\subsection{Textual Representations}
\us{To generate textual representations for media}, 
we collect data from \emph{Articles} and \emph{Wikipedia} pages. For \emph{Articles}, we extracted media details from MBFC, parsed news media pages with article links. 

Also, we retrieved article texts using custom scripts. Post-processing ensured that data were formatted into JSON. For \emph{Wikipedia}, we gather media outlet information by extracting HTML from \emph{Wikipedia} pages and converting it to JSON. 

Furthermore, the goal of collecting \emph{Articles} was \emph{to capture what was written by the media outlet}, whereas \emph{Wikipedia} pages were collected \emph{to reflect what was written about the outlet}. Together, these sources were used to assess the media's current positioning, standing, or ideological leaning.
Detailed data collection steps are provided in Appendix~\ref{sec:appendix}.

\begin{table}[!t]
  \centering
\resizebox{0.47\textwidth}{!}{%
    \begin{tabular}{c|l|r|c|r|r}
       \toprule
       \multirow{2}{*}{\textbf{Dataset}} & \multirow{2}{*}{\textbf{Representation}} & \multirow{2}{*}{\textbf{\# Sampled Label}} & \multirow{2}{*}{\textbf{\# of Levels}} & \multirow{2}{*}{\textbf{\# of Nodes}} & \multirow{2}{*}{\textbf{\# of Edges}} \\
       & & & & & \\
       \midrule
       \multirow{5}{*}{\textbf{ACL-2020}} 
       & Alexa graph            & 475   & 3 & 67,350 & 100,261 \\
       & LLM-graph          & 475   & 3 & 18,508 & 43,410 \\
       & Hyperlink graph     & 475   & 3 & 88,722 & 169,363 \\
       & Wikipedia        & 475   & - & - & - \\
       & Media (Articles) & 475   & - & - & - \\
       \midrule
       \multirow{5}{*}{\textbf{MBFC-2025}} 
       & Alexa graph           & 2,550 & 3 & 67,350 & 100,261 \\
       & LLM-graph      & 2,550 & 3 & 80,385 & 194,755 \\
       & Hyperlink graph   & 2,550 & 2 & 528,214 & 1,994,492 \\
       & Wikipedia        & 2,550 & - & - & - \\
       & Media (Articles) & 2,550 & - & - & - \\
       \bottomrule
    \end{tabular}
    }
    \caption{Statistics of representations, sampled labels, and graph structures for ACL-2020 and MBFC-2025.}
    \label{tab:combined-stats}
\end{table}

\begin{table}[!t]
\centering
\resizebox{0.5\textwidth}{!}{%
\begin{tabular}{l|c|c|c|c|c}
\toprule
\textbf{Dataset} & \textbf{Task} & \textbf{Train} & \textbf{Dev} & \textbf{Test} & \textbf{Total} \\
\midrule
\textbf{MBFC-2025} & Political Bias \& Factuality & 2040 & 256 & 254 & 2550 \\
\midrule
\textbf{ACL-2020} & Political Bias \& Factuality & 383  & 43  & 49  & 475 \\
\bottomrule
\end{tabular}
}
\caption{Statistics for train, development, and test sets.}
\label{splits}
\end{table}

\subsection{Datasets Statistics}
\label{datastat}

Table~\ref{label distribution} presents the label distributions for political bias and factuality. Data were split into train/validation/test sets using an 80/10/10 ratio without overlap of media outlets or articles on the same story and event to prevent data leakage. Stratified sampling ensured balanced class distributions across splits. Table~\ref{tab:combined-stats} summarizes the representation types, sampled labels, and graph structures, while Table~\ref{splits} reports detailed statistics for each split in ACL-2020 and MBFC-2025 datasets.

\section{Methodology}

\subsection{Problem Formulation}

\blue{We formulate news media profiling as a multi-modal (refers to the number of representations) classification problem, where political bias and factuality are predicted independently using various representations.} Given a media outlet $F_i$, we represent it with a set of five representations:
\[
\mathcal{F}_i = \{F^{(a)}, F^{(h)}, F^{(l)}\, F^{(t)}, F^{(w)}\}
\]
\vspace{-0.2em}
\us{Where $F^{(a)}$ denotes the Alexa graph, $F^{(h)}$ the Hyperlink graph, $F^{(l)}$ the LLM generated graph, $F^{(t)}$ the media articles, and $F^{(w)}$ the Wikipedia descriptions.}

\us{Let $\mathcal{G}: \mathcal{F} \rightarrow y$ denote a prediction function mapping the multi-view representations to a target label $y$, corresponds to either political bias or factuality. Our objective is to identify effective combinations of representations and fusion approaches that maximize predictive performance for each task.}



\subsection{Multi-View Integration}
\begin{figure}[t!]
    \includegraphics[width=1.0\linewidth]{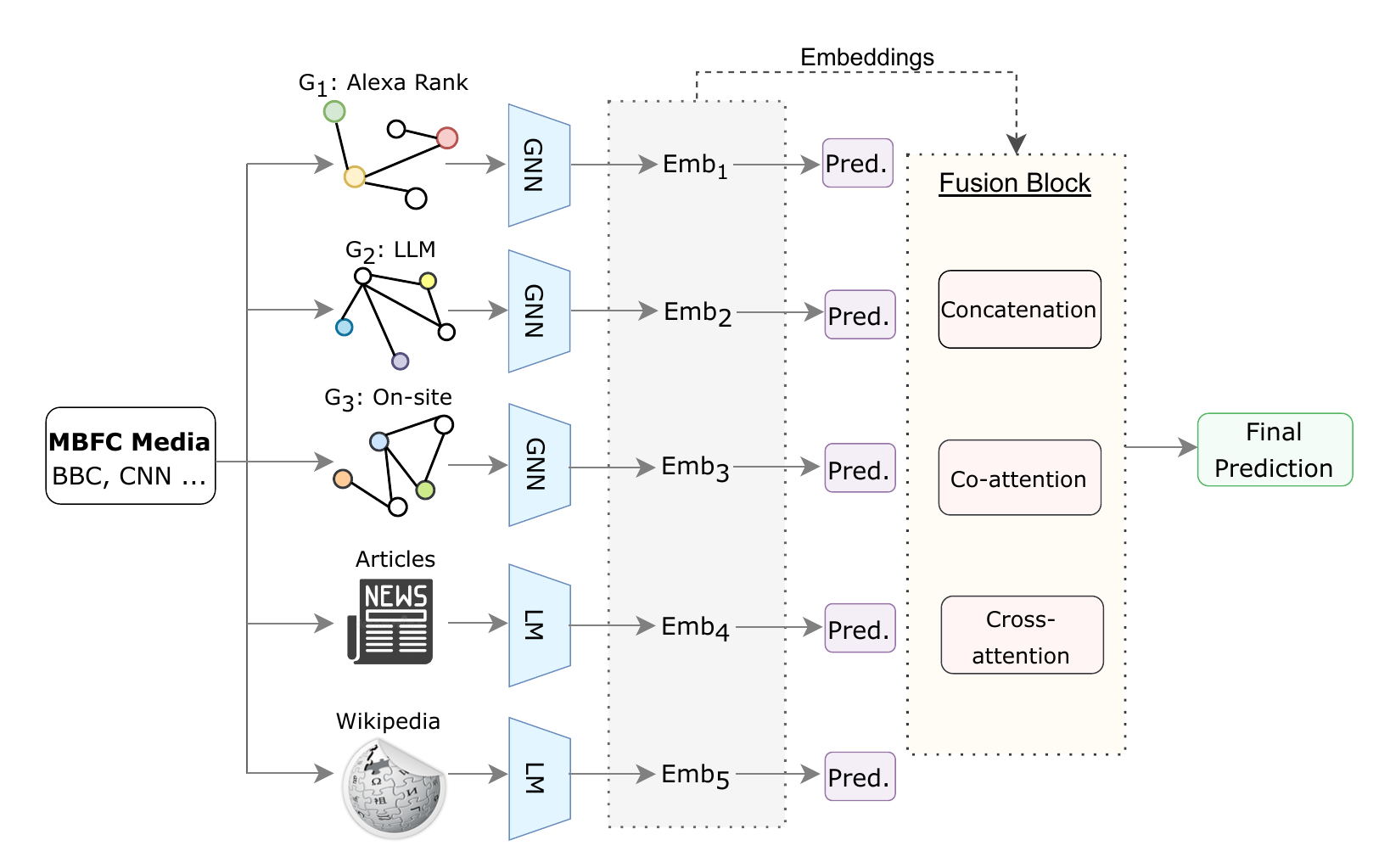}
    \caption{
    \textbf{End-to-end pipeline of our approach.} Given an MBFC media outlet, we construct multiple graphs and textual views. GNNs and pre-trained language models (PLMs) generate outlet-level embeddings for each view, followed by view-specific predictions. The embeddings are combined through various fusion mechanisms.    }
    \label{Mainfig}
\end{figure}

In this section, we present our multi-view integration framework, shown in Figure~\ref{Mainfig}. For each target medium, we have a variety of representations, such as (\emph{i})~\emph{Alexa graph}, (\emph{ii})~\emph{Hyperlink graph}, (\emph{iii})~\emph{LLM-graph}, (\emph{iv})~\emph{Articles}, and (\emph{v})~\emph{Wikipedia}. Textual representations are extracted using advanced text models, while recent GNNs effectively handle graph representations. These are then fused using several complementary fusion strategies to provide a comprehensive view of the website's content and structure, and to find \us{a better} combination of representations.

\subsubsection{GNN Models}
In our study, we use three GNNs: Residual Gated Graph Convolutional Networks (ResGatedGCNs) \cite{bresson2017residual}, Graph Convolutional Networks (GraphConv) \cite{morris2019weisfeiler}, and GraphSAGE \cite{hamilton2017inductive}. GraphConv extends convolutional operations to graphs, enabling the aggregation of information from neighboring nodes to learn node representations. 

GraphSAGE introduces a sampling method that selects a fixed-size neighborhood for each node during training, reducing memory usage and computational costs while improving scalability for large graphs. ResGatedGCNs combine residual connections with gated mechanisms to mitigate over-smoothing, allowing for more stable information flow across layers.

\subsubsection{Textual Models}

We evaluate a diverse range of textual models, including: classical SVM with TF-IDF~\cite{hearst1998support, sparck1972statistical}, BERT\textsubscript{Base}~\cite{devlin-etal-2019-bert}, RoBERTa\textsubscript{Base}~\cite{liu2019roberta}, DistilBERT\textsubscript{Base}~\cite{sanh2019distilbert}, and ALBERT\textsubscript{Base}~\cite{lan2019albert}, Mistral 7B~\cite{jiang2023mistral}, Llama-2 7B~\cite{touvron2023llama}, and GPT-4o~\cite{achiam2023gpt}.





\textbf{Textual representations:} To obtain outlet-level labels, we aggregate individual article predictions using hard and soft voting. This ensemble approach \citep{DBLP:journals/jcss/FreundS97} helps reduce variance and ensures robust classification. We also used each outlet’s Wikipedia description as an additional signal, capturing broader contextual and historical information. To leverage complementary information from both sources, we concatenated article and Wikipedia embeddings and applied the same aggregation strategies used for media articles.

\subsubsection{Representations Fusion Approaches}
We use DistilBERT\textsubscript{Base}, chosen for efficiency, to generate embeddings from media articles and Wikipedia descriptions. To complement textual representations, we use GNNs to learn structural embeddings that capture relational patterns. We then combine textual and structural embeddings using fusion strategies. \emph{(i) SVM Fusion:} Following \citet{baly-etal-2020-written}, we concatenate embeddings and apply an SVM classifier. \emph{(ii) MLP Fusion:} We apply a single-layer perceptron to combine embeddings, as done by \citet{mlpfusion}, since this method outperforms simpler operations such as addition or multiplication. \emph{(iii) Self-Attention Fusion:} We apply self-attention to the concatenated embeddings to capture feature importance. \emph{(iv) Late Attention Fusion:} Namely, \emph{(a) Cross-Attention}, where one representation attends to the other, and \emph{(b) Co-Attention}, where both attend to each other. \us{\emph{(v) RL-based Dynamic Fusion:} To overcome static fusion limitations, we propose an RL-based method that learns outlet-specific weights for each view.}

\paragraph{RL-based Dynamic Fusion.}
We cast fusion as a contextual bandit where actions do not affect state transitions. The state concatenates the multi-view embeddings for outlet $F_i$,
$s_t=\{F^{(a)},F^{(h)},F^{(l)},F^{(t)},F^{(w)}\}$.
The action is a continuous weight vector $\mathbf{w}\in\mathbb{R}^5$ with $w_k\in[0,1]$ indicating the importance of each view. We compute the fused embedding as
\[
E_{\text{fused}}=\sum_{k=1}^5 w_k\,F^{(k)},
\]
and define the reward as the probability of the true label under a fixed classifier,
\[
r_t = P\!\left(y_{\text{true}} \mid E_{\text{fused}}\right).
\]
The goal is to learn a policy $\pi(a\mid s)$ that maximizes the immediate expected reward. We implement the environment in Gymnasium~\cite{DBLP:journals/corr/abs-2407-17032} and train the policy with Proximal Policy Optimization (PPO)~\cite{schulman2017proximal} in Stable-Baselines3~\cite{stable-baselines3}, setting the discount factor near zero to emphasize immediate rewards.

\section{Experiments and Evaluation}

\begin{table}[!t]
    \centering
    \resizebox{0.5\textwidth}{!}{
    \begin{tabular}{l|cccc|cccc}
        \toprule
        \textbf{Hyper-parameter} & \multicolumn{4}{c|}{\textbf{ACL-2020}} & \multicolumn{4}{c}{\textbf{MBFC-2025}} \\
        \midrule
        & \textbf{BERT} & \textbf{RoBERTa} & \textbf{DistilBERT} & \textbf{ALBERT} & \textbf{BERT} & \textbf{RoBERTa} & \textbf{DistilBERT} & \textbf{ALBERT} \\
        \midrule
        Batch size & 80 & 95 & 110 & 120 & 80 & 90 & 100 & 100 \\
        Max length & 256 & 256 & 512 & 512 & 128 & 128 & 256 & 256 \\
        Epochs & 4 & 4 & 5 & 6 & 5 & 4 & 6 & 5 \\
        Learning rate & 2e-5 & 2e-5 & 2e-5 & 2e-5 & 2e-5 & 2e-5 & 2e-5 & 2e-5 \\
        \bottomrule
    \end{tabular}
    }
    \caption{Experimental setup for PLMs on ACL-2020 and MBFC-2025 datasets.}
    \label{hyperpar_combined}
\end{table}

\subsection{Experimental Setup}

\textbf{GNNs.} We learn media graph embeddings using a contrastive objective in an unsupervised setting, where labeled nodes constitute under 1\%. 
For each graph, we obtain 64-dimensional embeddings using GCN, GraphSAGE, and ResGatedGCN. All models share the same hyperparameters: \emph{epochs} = 50, \emph{layers} = 4, \emph{hidden size} = 128, \emph{batch size} = 128, \emph{learning rate} = 1e-4, \emph{dropout} = 0.5.

\noindent\textbf{PLMs and SVM.} We use consistent hyperparameters across tasks, with minor adjustments for PLMs detailed in Table~\ref{hyperpar_combined}. For SVM, we use a linear kernel with a maximum of 60 iterations and a tolerance of 0.01 on both datasets.

\noindent\textbf{LLM Setup.} Due to input length constraints, articles are first summarized using BART~\cite{lewis2019bart} to 250-300 words. For each outlet, we select five articles that focus on political, economic, and social topics. These summaries are then used as inputs to LLMs through task-specific prompts. We sample 100 outlets from ACL-2020 and 300 from MBFC-2025 for zero- and few-shot (1, 3, 5) evaluation. For few-shot settings, examples are selected to ensure label diversity. 

The final outlet-level predictions are obtained using both soft and hard voting on five article-level predictions. Summarization provides additional context, while voting reduces variance and bias, resulting in more stable media-level classification.

\label{prompt}

\begin{lstlisting}[language=Python,frame=lines,framesep=2mm,basicstyle=\ttfamily\footnotesize\linespread{1.2}\selectfont,mathescape=true,breaklines=true,showstringspaces=false]
system_prompt = '''Summarize the following news article in 250-300 words. Ensure the summary covers the key points and main details.'''
user_prompt = "{article}"
\end{lstlisting}

The summarized article (replacing \textbf{\{article\}}) is used as evidence for downstream classification.

\begin{lstlisting}[language=Python,frame=lines,framesep=1mm,basicstyle=\ttfamily\footnotesize\linespread{1.2}\selectfont,mathescape=true,breaklines=true,showstringspaces=false]
system_prompt = '''You are an expert in media analysis. Classify the factual reporting level of the given news article from {media} into one of the following:
- high
- mixed
- low
Return -1 if uncertain.'''
user_prompt = "{article}"
\end{lstlisting}

\begin{lstlisting}[language=Python,frame=lines,framesep=1mm,basicstyle=\ttfamily\footnotesize\linespread{1}\selectfont,mathescape=true,breaklines=true,showstringspaces=false]
system_prompt = '''You are an expert in media analysis. Classify the political bias of the given news article from {media} into one of the following:
- left
- center
- right
Return -1 if uncertain.'''
user_prompt = "{article}"
\end{lstlisting}

We replace \textbf{\{media\}} placeholder with the outlet name, and \textbf{\{article\}} with the summarized article. 
For MBFC-2025, we extend the label space to include finer-grained categories: \textit{very high} and \textit{very low} for factuality, and \textit{left-center} and \textit{right-center} for political bias.

\definecolor{BestGreen}{RGB}{220,245,220}
\definecolor{LowBlue}{RGB}{220,230,250}

\definecolor{BestGreenText}{RGB}{0,120,0}
\definecolor{LowBlueText}{RGB}{0,70,160}

\begin{table*}[!t]
\centering
\resizebox{1.0\textwidth}{!}{
\large
\begin{tabular}{l|c|ccccc}
\toprule
\textbf{Models} & \textbf{Representation} & \textbf{MAE} & \textbf{Macro-F1} & \textbf{Accuracy} & \textbf{Precision} & \textbf{Recall} \\
\midrule
\multicolumn{7}{c}{\textbf{ACL-2020}} \\
\midrule

GREENER~\cite{EMNLP2022:GREENER} & $Twitter$ + $YouTube$ + $Alexa$ + $A$ + $W$  & - & 91.93 & 92.08 & - & - \\
MGM~\cite{manzoor2025mgm} & $A + Graph$ $Embeddings$  & - & 93.08 & 93.45 & - & 93.19 \\

\midrule
Majority class &  & \cellcolor{LowBlue}0.633 & \cellcolor{LowBlue}17.91 & \cellcolor{LowBlue}36.73 & \cellcolor{LowBlue}12.24 & \cellcolor{LowBlue}33.33 \\
RoBERTa$_{\text{Base}}$ Soft Voting & $A$ & 0.285 & 76.16 & 77.55 & 80.51 & 75.91  \\
RoBERTa$_{\text{Base}}$ & $W$ & 0.484 & 66.33 & 70.47 & 66.81 & 66.38 \\
RoBERTa$_{\text{Base}}$ Soft Voting & $A$ + $W$ & 0.224 & 81.47 & 81.63 & 83.61 & 81.20 \\

\midrule
SVM & $Alexa_{ResGGCN}$ & 0.596 & 60.54 & 61.70 & 60.67 & 60.78 \\
SVM & $Alexa_{ResGGCN} + A$ & 0.106 & 91.56 & 91.49 & 91.61 & 91.61 \\
Co-attention & $Alexa_{GCN}$ + $Hyperlink_{SAGE}$ + $A$ & 0.111 & 93.20 & 93.33 & 94.12 & 93.28 \\
SVM & $Alexa_{ResGGCN}$ + $Hyperlink_{SAGE}$ + $A$ + $W$ & \cellcolor{BestGreen}\textbf{0.067} & 95.53  & 95.56 & 95.83 & 95.66 \\
SVM &  $LLM_{GCN}$ + $Alexa_{ResGGCN}$ + $Hyperlink_{SAGE}$ + $A$ + $W$ & 0.089 & 93.46 & 93.33 & 94.12 & 93.70 \\
RL (PPO) & $Alexa_{GCN}$ + $Hyperlink_{SAGE}$ + $LLM_{GCN}$ + $A$ + $W$ & 0.098 & \cellcolor{BestGreen}\textbf{97.60} & \cellcolor{BestGreen}\textbf{97.60} & \cellcolor{BestGreen}\textbf{97.30} & \cellcolor{BestGreen}\textbf{98.00} \\

\midrule
\multicolumn{7}{c}{\textbf{MBFC-2025}} \\
\midrule

Majority class &  & \cellcolor{LowBlue}1.071 & \cellcolor{LowBlue}9.21 & \cellcolor{LowBlue}29.92 & \cellcolor{LowBlue}5.98 & \cellcolor{LowBlue}20.00 \\
ALBERT$_{\text{Base}}$ Soft Voting & $A$ & 0.457 & 71.74 & 74.41 & 71.84 & 71.61 \\
Ensemble Hard Voting & $W$ & 0.623 & 63.19 & 64.15 & 65.33 & 62.86 \\
Ensemble Soft Voting & $A$ + $W$ & 0.425 & 72.41 & 75.47 & 73.69 & 72.36 \\

\midrule
SVM & $Alexa_{GCN}$ & 1.010 & 40.44 & 46.08 & 42.59 & 41.94 \\
SVM & $A$ + $W$ & 0.405 & 69.09 & 73.14 & 71.00 & 68.67 \\
SVM & $Hyperlink_{GCN}$ + $A$ + $W$ & \cellcolor{BestGreen}\textbf{0.395} & 69.29 & 73.18 & 70.81 & 69.05 \\
MLP & $LLM_{GCN}$ + $Hyperlink_{SAGE}$ + $A$ + $W$ & 0.482 & 67.04 & 70.91 & 66.93 & 68.46 \\
Self-attention & $LLM_{SAGE}$ + $LLM_{GCN}$ + $Hyperlink_{SAGE}$ + $A$ + $W$ & 0.486 & 67.79 & 70.00 & 68.53 & 71.09 \\
RL (PPO) & $Alexa_{GCN}$ + $Hyperlink_{SAGE}$ + $LLM_{GCN}$ + $A$ + $W$ & 0.551 & \cellcolor{BestGreen}\textbf{78.00} & \cellcolor{BestGreen}\textbf{77.90} & \cellcolor{BestGreen}\textbf{77.80} & \cellcolor{BestGreen}\textbf{78.30} \\

\bottomrule
\end{tabular}
}

\caption{Evaluation results using multiple representations for \emph{political bias} detection and baselines from prior work. The representation column indicates the embeddings used as input: Articles ($A$), Wikipedia ($W$), Alexa graph ($Alexa$), Hyperlink graph ($Hyperlink$), and LLM-based graph ($LLM$). Subscripts denote the GNN encoder. \textcolor{BestGreenText}{\textbf{Green}} marks the best scores, while \textcolor{LowBlueText}{\textbf{Blue}} marks the lowest scores.}
\label{political_bias_results}
\end{table*}

\noindent\textbf{RL Agent.} We implemented the RL agent using PPO. To model the problem as a contextual bandit, we set the discount factor to $0$, ensuring optimization focuses on immediate rewards. The policy network is an MLP with two hidden layers of 128 neurons and \texttt{tanh} activations. We use a learning rate of $1\times10^{-4}$, a batch size of 256, and a rollout size of 1024 environment steps per policy update.

\textbf{Evaluation Measures.} We use MAE, Macro-F1, Accuracy, Precision, and Recall, with MAE reflecting the ordinal nature of labels~\cite{baly2018predicting, baly-etal-2020-detect}. Hyperparameters for GNNs and PLMs are tuned on development sets using common values to balance performance and efficiency. All experiments run on an NVIDIA RTX A6000 48GB GPU.

\subsection{Evaluation}

We discuss our best results in \S~\ref{res}, with additional results in Appendix~\ref{results}. The model analysis is provided in \S~\ref{discussion}, and key findings are summarized in \S~\ref{findings}.

\begin{table*}[!t]
\centering
\resizebox{1.0\textwidth}{!}{
\large
\begin{tabular}{l|c|ccccc}
\toprule
\textbf{Models} & \textbf{Representation} & \textbf{MAE} & \textbf{Macro-F1} & \textbf{Accuracy} & \textbf{Precision} & \textbf{Recall} \\
\midrule

\multicolumn{7}{c}{\textbf{ACL-2020}} \\
\midrule

Node classification (NC)~\cite{mehta2022tackling} &  $Twitter$ + $YouTube$ + $Social Graph$ + $Users Profiles$  + $A$ & - & 68.90 & 63.72 & - & - \\
Inf\textsc{Op} Best Model~\cite{mehta2022tackling} & $Twitter$ + $YouTube$ + $Social Graph$ + $Users Profiles$  + $A$ & - & 72.55 & 66.89 & - & - \\
GREENER~\cite{EMNLP2022:GREENER} & $Twitter$ + $YouTube$ + $Alexa$ + $A$ + $W$  & - & 69.61 & 74.27 & - & - \\
MGM~\cite{manzoor2025mgm} & $A + Graph$ $Embeddings$  & - & \cellcolor{BestGreen}\textbf{79.72} & 84.21 & - & 76.54 \\

\midrule
Majority class & & 0.571 & \cellcolor{LowBlue}24.79 & \cellcolor{LowBlue}59.18 & \cellcolor{LowBlue}19.73 & \cellcolor{LowBlue}33.33 \\

ALBERT$_{\text{Base}}$ Hard Voting & $A$ & 0.449 & 51.46 & 67.35 & 76.14 & 50.72 \\
SVM$_{\text{TF-IDF}}$ & $W$ & 0.551 & 36.92 & 57.14 & 34.12 & 40.33 \\
BERT$_{\text{Base}}$ & $W$ & 0.531 & 30.20 & 61.22 & 37.23 & 36.11 \\
ALBERT$_{\text{Base}}$ Hard Voting &  $A + W$ & 0.469 & 42.42 & 65.31 & 77.04 & 43.06 \\

\midrule
SVM & $Alexa_{GCN}$ & \cellcolor{LowBlue}0.574 & 42.06 & 55.32 & 49.30 & 42.49 \\
Cross-attention & $Hyperlink_{GCN}$ + $A$ & \cellcolor{BestGreen}\textbf{0.178} & 78.63 & 84.44 & 81.25 & \cellcolor{BestGreen}\textbf{77.59} \\
MLP & $Alexa_{ResGGCN}$ + $Hyperlink_{SAGE}$ + $A$ & \cellcolor{BestGreen}\textbf{0.178} & 76.07 & 82.22 & 78.86 & 74.82 \\
Self-attention & $LLM_{GCN}$ + $Hyperlink_{SAGE}$ + $Hyperlink_{GCN}$ + $A$ & 0.267 & 73.40 & 77.78 & 74.60 & 74.73 \\
Cross-attention & $LLM_{GCN}$ + $Alexa_{GCN}$ + $Hyperlink_{SAGE}$ + $Hyperlink_{GCN}$ + $A$ & 0.222 & 74.63 & 80.00 & 75.79 & 76.01 \\
RL (PPO) & $Alexa_{GCN}$ + $Hyperlink_{SAGE}$ + $LLM_{GCN}$ + $A$ + $W$ & 0.183 & 76.80 & \cellcolor{BestGreen}\textbf{86.60} & \cellcolor{BestGreen}\textbf{91.20} & 71.70 \\

\midrule
\multicolumn{7}{c}{\textbf{MBFC-2025}} \\
\midrule

Majority class &  & \cellcolor{LowBlue}0.567 & \cellcolor{LowBlue}13.81 & \cellcolor{LowBlue}52.76 & \cellcolor{LowBlue}10.55 & \cellcolor{LowBlue}20.00 \\
ALBERT$_{\text{Base}}$ Hard Voting  & $A$ & 0.457 & \cellcolor{BestGreen}\textbf{71.74} & \cellcolor{BestGreen}\textbf{74.41} & \cellcolor{BestGreen}\textbf{71.84} & \cellcolor{BestGreen}\textbf{71.61} \\

RoBERTa$_{\text{Base}}$ & $W$ & 0.413 & 29.49 & 63.68 & 45.47 & 30.04 \\
DistilBERT$_{\text{Base}}$ & $W$ & 0.398 & 27.45 & 64.68 & 26.07 & 29.18 \\

RoBERTa$_{\text{Base}}$ Hard Voting & $A + W$ & 0.313 & 41.58 & 71.64 & 68.58 & 38.63 \\

\midrule
SVM & $Alexa_{GCN}$ & 0.405 & 30.76 & 65.29 & 44.33 & 29.94 \\
Cross-attention & $A$ + $W$ & \cellcolor{BestGreen}\textbf{0.277} & 51.78 & 75.62 & 49.33 & 58.98 \\
Co-attention & $Alexa_{ResGGCN}$ + $A$ + $W$ & 0.291 & 56.94 & 74.36 & 55.06 & 61.74 \\
Cross-attention &  $LLM_{GCN}$ + $Hyperlink_{SAGE}$ + $A$ + $W$ & 0.306 & 50.82 & 71.76 & 51.35 & 61.72 \\
Cross-attention & $LLM_{SAGE}$ + $LLM_{GCN}$ + $Hyperlink_{GCN}$ + $A$ + $W$ & 0.310 & 53.30 & 71.76 & 50.90 & 64.32 \\
RL (PPO) & $Alexa_{GCN}$ + $Hyperlink_{SAGE}$ + $LLM_{GCN}$ + $A$ + $W$ & 0.249 & 55.30 & 72.10 & 67.10 & 62.80 \\

\bottomrule
\end{tabular}
}

\caption{Evaluation results using multiple representations for \emph{factuality} detection and baselines from prior work. Representations include Articles ($A$), Wikipedia ($W$), Alexa graph ($Alexa$), Hyperlink graph ($Hyperlink$), and LLM-based graph ($LLM$). Subscripts denote the GNN encoder. \textcolor{BestGreenText}{\textbf{Green}} marks the best scores, while \textcolor{LowBlueText}{\textbf{Blue}} marks the lowest scores.}
\label{fact_results}
\end{table*}

\subsubsection{Results}
\label{res}

\us{As illustrated in Table~\ref{political_bias_results} for \emph{political bias}, the combination of media Articles and Wikipedia representations, together with Alexa graph (GCN) and Hyperlink graph (GraphSAGE), using an SVM achieved the strongest results among all static baselines on ACL-2020, with a Macro-F1 of 95.53\%.} 

Moreover, our RL-based fusion strategy consistently outperformed all static approaches and achieved a higher Macro-F1 of 97.60\%. On the larger and more challenging MBFC-2025, our RL-based dynamic fusion strategy again significantly outperformed all static baselines and ensemble methods. 

Although the best static fusion achieved a Macro-F1 of 69.29\% and ensemble voting reached 72.41\%, our RL agent achieved a new state-of-the-art Macro-F1 of \textbf{78.00\%}. 
These results demonstrate the efficacy of dynamic weight assignment, particularly in complex, multi-view settings where static fusion strategies performed poorly. Detailed results of \emph{factuality} prediction among both datasets are given in Table~\ref{fact_results}.

In more details, for \emph{factuality} prediction, our RL-baseline fusion further achieved accuracy to 86.60\% while maintaining a comparable Macro-F1 of 76.80\%. On MBFC-2025, the RL agent achieved 55.30\% Macro-F1, performing on par with other fusion techniques such as Co-attention (56.94\%). However, text-only models remained dominant for the factuality task overall, with ALBERT\textsubscript{Base} reaching 71.74\% Macro-F1 through hard voting.

\subsubsection{GNN Embeddings for SVM}
\label{appx:gnn_svm}
For \emph{factuality} (Table~\ref{factuality_results_gnn}), in ACL-2020 the Alexa graph with GCN achieved the best Macro-F1 (42.06\%), while ResGatedGCN obtained the lowest MAE (0.532). In MBFC-2025, LLM-based embeddings with GraphSAGE achieved the lowest MAE (0.365), and the Alexa graph with GCN yielded the best Macro-F1 (30.76\%), demonstrating strong performance across evaluation settings.

For \emph{political bias} (Table~\ref{bias_results_gnn}), in ACL-2020 the Alexa graph with ResGatedGCN achieved both the best Macro-F1 (60.54\%) and MAE (0.596). In MBFC-2025, the Alexa graph with GCN gave the highest Macro-F1 (40.44\%), while LLM-based embeddings with GCN achieved the lowest MAE (0.787).

\subsubsection{Discussion}
\label{discussion}

As shown in Tables~\ref{political_bias_results} and \ref{fact_results}, we evaluated combinations of representations to detect \emph{factuality} and \emph{political bias} in various methods and datasets. 

\noindent \textbf{GNNs.} 
As detailed in Section~\ref{appx:gnn_svm}, there is no particular trend in GNN performance. For ACL-2020, SVM+Alexa rank features yield the best factuality prediction results; for MBFC-2025, LLM+GraphSAGE yields the lowest MAE, and Alexa+GCN yields the best Macro-F1. For political bias in the ACL dataset, the best combination was SVM+Alexa + ResGatedGCN, while for MBFC-2025, Alexa+GCN gave the best Macro-F1, and LLM+GraphSAGE gave the lowest MAE 0.203.

\noindent \textbf{PLMs and LLMs.} PLMs and LLMs generally predict political bias better than factuality, with Wikipedia pages having a stronger influence on LLMs. Fine-tuned PLMs on both media articles and Wikipedia descriptions outperform those trained on Wikipedia alone. However, combining representations in LLMs reduces performance compared to a single representation.


\subsubsection{Empirical Findings}
\label{findings}

(\textit{i}) Political bias is easier to detect than factuality, as it is more explicit in language, while factuality requires deeper contextual understanding and implicit verification. It is also consistent with ~\citet{EMNLP2022:GREENER}.
(\textit{ii}) Single-view graph representations are suboptimal and combining multiple views yields stronger results by capturing diverse network dependencies. 

(\textit{iii}) PLMs benefit from multiple representations, whereas LLMs often perform best with a single high-quality view, likely because additional views introduce noise and increase optimization complexity in practice. (\textit{iv}) Multi-view evaluation improves predictive performance and interpretability by clarifying which representations and interactions matter most. However, as shown in Appendix~\ref{dependency}, adding more representations does not always improve the results, often leading to diminishing returns across different experimental settings.
(\textit{v}) The Alexa graph is the most effective view due to its broad coverage and informative node features. Hyperlink graphs provide complementary signals, whereas the LLM graph is weakest due to missing explicit node features and limited structural information.

(\textit{vi}) Performance is higher on ACL-2020 than on MBFC-2025, likely because the 5-point scale of MBFC introduces finer distinctions and greater ambiguity.
(\textit{vii}) RL-based fusion improves political bias detection, but not strong factuality prediction, as political bias signals are more consistent between views and support more reliable reward-driven selection. In contrast, factuality is more instance-dependent and requires stronger grounding, leading to noisier rewards and a policy more susceptible to spurious correlations. 
Overall, performance depends on task and representation: multi-view helps with stable signals, while factuality needs stronger grounding and is more sensitive.

\begin{table}[!t]
\centering
\resizebox{0.49\textwidth}{!}{
\large 
\begin{tabular}{l|ccccc}
\toprule
\textbf{Representation} & \textbf{MAE} & \textbf{Macro-F1} & \textbf{Accuracy} & \textbf{Precision} & \textbf{Recall} \\
\midrule
 & & \textbf{ACL-2020} & & & \\
\midrule
$LLM_{SAGE}$ & 0.596 & 27.64 & 55.32 & 35.19 & 34.43 \\
$LLM_{GCN}$ & 0.596 & 24.07 & 55.32 & 18.84 & 33.33 \\
$Alexa_{ResGGCN}$ & \textbf{0.532} & 40.53 & \textbf{57.45} & 46.12 & 41.39 \\
$Alexa_{GCN}$ & 0.574 & \textbf{42.06} & 55.32 & \textbf{49.30} & \textbf{42.49} \\
$Hyperlink_{SAGE}$ & 0.689 & 37.70 & 51.11 & 41.25 & 37.45 \\
$Hyperlink_{GCN}$ & 0.600 & 28.33 & 55.56 & 30.62 & 33.55 \\
\midrule
 & & \textbf{MBFC-2025} & & & \\
\midrule
$LLM_{SAGE}$ & \textbf{0.365} & 27.99 & \textbf{68.25} & 27.00 & 29.28 \\
$LLM_{GCN}$ & 0.373 & 28.03 & 67.46 & 26.80 & 29.39 \\
$Alexa_{ResGGCN}$ & 0.438 & 26.42 & 62.81 & 27.79 & 27.34 \\
$Alexa_{GCN}$ & 0.405 & \textbf{30.76} & 65.29 & \textbf{44.33} & \textbf{29.94} \\
$Hyperlink_{SAGE}$ & 0.489 & 23.33 & 59.39 & 23.21 & 23.93 \\
$Hyperlink_{GCN}$ & 0.476 & 22.12 & 61.57 & 23.65 & 23.60 \\
\bottomrule
\end{tabular}
}
\caption{Evaluation results using SVM with different GNN embeddings for \emph{factuality} detection across ACL-2020 and MBFC-2025 datasets.}
\label{factuality_results_gnn}
\end{table}

\section{Conclusion and Future Work}
We introduced \textbf{MBFC-2025}, a large-scale, fine-grained annotation set, and construct multi-view representations across two benchmarks. Through an extensive evaluation of embeddings and fusion strategies, including an RL-based dynamic variant, we achieve state-of-the-art results in ACL-2020 and establish strong benchmarks in MBFC-2025, where the RL agent outperforms static fusion, particularly for the detection of political bias. 

In future work, we plan to expand this into a multilingual corpus capturing bias across cultures, reducing the current U.S.-centric focus. Also, we plan to scale our modeling approach by incorporating graph neural networks (GNNs) and parameter-efficient fine-tuning of large language models, using methods such as LoRA~\cite{hu2021lora}, LoFT~\cite{tastan2025loft}, and potentially mixture-of-expert architectures~\cite{tastan2026mose}.

\begin{table}[!t]
\centering
\resizebox{0.49\textwidth}{!}{
\large 
\begin{tabular}{l|ccccc}
\toprule
\textbf{Representation} & \textbf{MAE} & \textbf{Macro-F1} & \textbf{Accuracy} & \textbf{Precision} & \textbf{Recall} \\
\midrule
 & & \textbf{ACL-2020} & & & \\
\midrule
$LLM_{SAGE}$ & 0.872 & 31.03 & 36.17 & 27.12 & 36.69 \\
$LLM_{GCN}$ & 0.809 & 35.14 & 40.43 & 59.87 & 40.74 \\
$Alexa_{ResGGCN}$ & \textbf{0.596} & \textbf{60.54} & \textbf{61.70} & \textbf{60.67} & \textbf{60.78} \\
$Alexa_{GCN}$ & \textbf{0.596} & 57.50 & 57.45 & 57.64 & 57.70 \\
$Hyperlink_{SAGE}$ & 0.911 & 23.92 & 24.44 & 23.93 & 24.09 \\
$Hyperlink_{GCN}$ & 0.667 & 41.49 & 42.22 & 45.42 & 41.46 \\
\midrule
 & & \textbf{MBFC-2025} & & & \\
\midrule
$LLM_{SAGE}$ & 0.802 & 26.77 & 43.48 & 28.97 & 31.56 \\
$LLM_{GCN}$ & \textbf{0.787} & 25.39 & 44.27 & 25.94 & 31.48 \\
$Alexa_{ResGGCN}$ & 1.088 & 38.61 & 43.14 & 38.45 & 39.54 \\
$Alexa_{GCN}$ & 1.010 & \textbf{40.44} & \textbf{46.08} & \textbf{42.59} & \textbf{41.94} \\
$Hyperlink_{SAGE}$ & 0.957 & 31.15 & 40.69 & 36.35 & 32.57 \\
$Hyperlink_{GCN}$ & 0.870 & 27.62 & 39.83 & 38.17 & 30.36 \\
\bottomrule
\end{tabular}
}
\caption{Evaluation results using SVM with different GNN embeddings for \emph{political bias} detection across ACL-2020 and MBFC-2025 datasets.}
\label{bias_results_gnn}
\end{table}

\subsection*{Limitations}

This study has data coverage limitations. Despite efforts to collect data from MBFC-annotated outlets, some websites could not be parsed. We also used Wikipedia for descriptive text and did not compare its labels with MBFC political leanings due to inconsistent availability and difficulty at scale. A further limitation concerns graph construction. The hyperlink and LLM graphs lack rich node attributes, so GNNs were trained using dummy features (e.g., a single integer per node). This likely constrained performance, although the embeddings still captured useful structural patterns.

Also, our experiments are limited by computational and modeling constraints. Due to time and memory requirements, we prioritized efficient models and did not explore fine-tuning, leaving it for future work. These constraints also prevented evaluation of larger LLMs. 

More broadly, our framework relies mainly on U.S.-centric political categories such as left, center, and right, which may not fully capture ideological variation in other cultural or political settings. The dataset should therefore not be used for fine-grained political bias or factuality detection, or for article-level analysis without broader context. Our work focuses on source-level political bias as a starting point, and future research should extend this setting to article- and claim-level prediction. Finally, while this study focuses on classification, it does not yet examine broader structural patterns in media ecosystems or more advanced fusion strategies, both of which remain important directions for future work.

\subsection*{Ethical Statement \& Bias}
Articles were collected according to legal and ethical standards, using publicly available content, adhering to site policies, and limiting access frequency to reduce server load. No paid or restricted content was accessed. Despite these safeguards, the data may still contain biases: Wikipedia-derived information can reflect implicit leanings, and both media sources and annotations may introduce systematic bias.
Although, we include a diverse range of outlets to mitigate these effects, the data may still underrepresent certain perspectives. As a result, findings should be interpreted with caution, particularly when generalizing beyond the sources studied or applying models in real-world settings.

Moreover, automated profiling and summarization may inadvertently reinforce polarization if users rely on condensed output rather than full context. Consequently, the framework is designed for transparency and research analysis, not for content filtering or personalization. 

To protect anonymity and data integrity, news articles and graph representations are not publicly available; instead, only the code, scraping recipes, and relevant URLs (including article/media links and labels) are provided to support reproducibility.

\bibliography{anthology,emnlp2023-latex/main}

\appendix

\clearpage


\section{Data Statement for MBFC-2025}
\label{sec:appendix}

\paragraph{Dataset Version} 1.0 (October 2024)

\paragraph{Data Statement} Version 1.0 (October 2024)

\paragraph{Data Collection Period} The dataset was collected from February to October 2024.

\paragraph{A.1 Executive Summary:}

MBFC-2025 is a dataset focused on analyzing political bias and factuality in English-language media at the outlet level. The dataset includes \emph{Articles} from various news media outlets, combined with \emph{Wikipedia} descriptions and graph-based representations. This dataset is designed to support research in media analysis.

\paragraph{A.2 Granularity:}

The dataset analysis is presented on a 5-point scale for political bias (Left, Left-Center, Center, Right-Center, Right) and factuality (Very High, High, Mixed, Low, Very Low).

\paragraph{A.3 Documentation for Source Datasets:}

The dataset was sourced from a variety of English-language media outlets, annotated by Media Bias/Fact Check experts. Data were collected through a systematic process that involved extraction of \emph{Articles} from news media websites and their corresponding \emph{Wikipedia} descriptions. Graph-based representations were obtained using the Alexa graph tool, Hyperlink graph, and LLM. The selection of sources was guided by relevance, credibility, and the objective of covering a broad spectrum of political bias and factuality.

\paragraph{A.4 Annotation Process:}

Political bias and factuality annotations were conducted by experts from Media Bias/Fact Check following a standardized protocol.

\paragraph{A.5 Intended Use:}

The dataset is intended for research on political bias and factuality at the media-level. It is particularly suited for tasks such as the detection of political bias and factuality, media analysis, and related computational studies.



\paragraph{A.7 Data Collection:}

\textbf{Label Collection from MBFC:} The labels were scraped from MBFC’s website using a web crawler, extracting political bias and factuality annotations for each media. 
To ensure accuracy, we parsed the HTML structure, identified classification tags, structured the data, and manually verified labels and media alignments to confirm correct extraction.

\textbf{Articles:} To collect articles, we focus on sections that cover political, economic, and social issues, selecting a variety of specific topics within these areas. The process involved two main steps: \emph{(i)} we exclusively used media sources annotated by experts from Media Bias/Fact Check (MBFC), and \emph{(ii)} we selected active media outlets. From these sources, we collected up to 30 front-page articles from each website.

The data collection process consisted of four stages: \emph{(i)} we gathered media sources from MBFC, extracting each source's details as JSON lines from the HTML code after manually verifying accessibility through its links, \emph{(ii)} we parsed front-page article links, excluding menu links and collecting only internal domain links over 65 characters, \emph{(iii)} the titles and article texts were retrieved using a combination of script automation and manual testing to ensure successful extraction, and \emph{(iv)} the post-processing stage involved formatting the data into the required JSON format.

\textbf{Wikipedia:} To locate the \emph{Wikipedia} link, we first searched for the name of the media outlet online. We ensure that the link points to a \emph{Wikipedia} entry about the media outlet. Next, we extracted the webpage in its uniform HTML format from the \emph{Wikipedia} website. Then we parsed the HTML to extract the page content.
The last step of the post-processing phase was to transform the gathered data into the necessary JSON format.

\section{Statistics of Alexa Media Graph}
\label{stat_of_alexa}

Table \ref{stat_rank} presents the statistics for the media graph constructed with the Alexa Rank tool based on audience overlap. The graph levels correspond to iterative expansions conducted by the authors \cite{EMNLP2022:GREENER} to enrich the graph and improve the contextual representation of the media within the network. The table clearly shows that as we progress to higher levels, both the number of nodes and edges increase, while the number of graph components decreases, which is beneficial for GNNs in learning effective node representations. However, at the highest level (Level 3), there are still 44 disconnected components, which may hinder the GNNs' ability to learn or generalize, potentially leading to sub-optimal node representations and lower performance on classification tasks.

\begin{table}[ht]
\centering
\resizebox{0.46\textwidth}{!}{
\begin{tabular}{lcccc}
\toprule
\textbf{Level} & \textbf{Nodes} & \textbf{Edges} & \textbf{\# of components} & \textbf{Avg. \# of nodes} \\ \midrule
0 & 4563 & 20210 & 326 & 10.7 \\ 
1 & 10161 & 28779 & 142 & 71.55 \\ 
2 & 26573 & 78600 & 75 & 354.3 \\ 
3 & 67351 & 200488 & 44 & 1530.7 \\ \bottomrule
\end{tabular}}
\caption{Graph statistics across different levels.}
\label{stat_rank}
\end{table}



    
    
    

\section{Results}
\label{results}

Our experimental results are shown in Tables \ref{baselines}-\ref{Task 3B}.
Table~\ref{baselines} shows the dummy-classifier baseline results for \emph{political bias} and \emph{factuality} on ACL-2020 and MBFC-2025. Table~\ref{LLM} shows the GPT-4o hard-voting results across the same two tasks and datasets. Table~\ref{Task 1A} shows the article-level \emph{political bias} classification results using hard and soft voting for PLMs, LLMs, and their ensembles. Table~\ref{Task 1B} shows the corresponding article-level \emph{factuality} classification results. Table~\ref{Task 2} shows the media-level results on Wikipedia-based datasets for both \emph{political bias} and \emph{factuality}, including independent frameworks and ensemble models under hard-voting and soft-voting settings. Finally, Table~\ref{Task 3A} and Table~\ref{Task 3B} show the \emph{political bias} and \emph{factuality} results, respectively, for the third experimental setting, again comparing PLMs, LLMs, and ensemble methods under both voting strategies.

\section{Dependency on Representations}
\label{dependency}

\begin{figure*}[t!]
    \centering
    \begin{minipage}{0.54\linewidth}
        \centering
        \includegraphics[width=\textwidth]{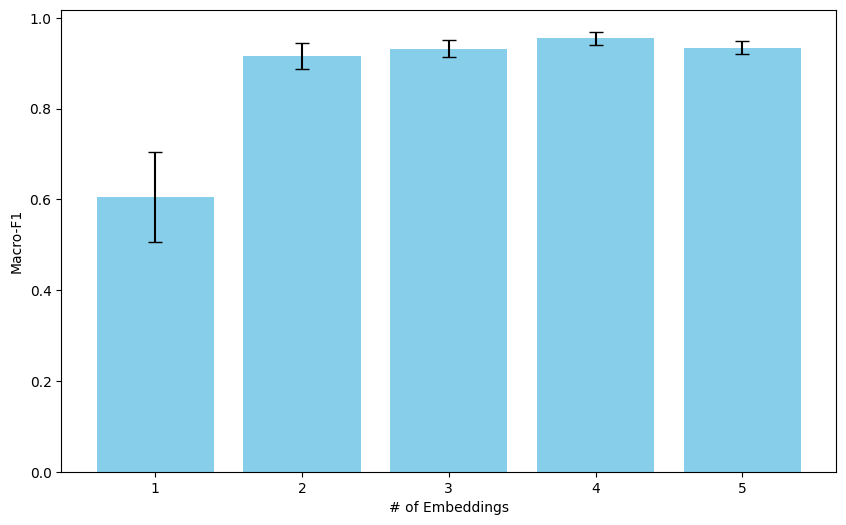}
        \subcaption{Political Bias - ACL2020.}
    \end{minipage}
    \hfill
    \begin{minipage}{0.54\linewidth}
        \centering
        \includegraphics[width=\textwidth]{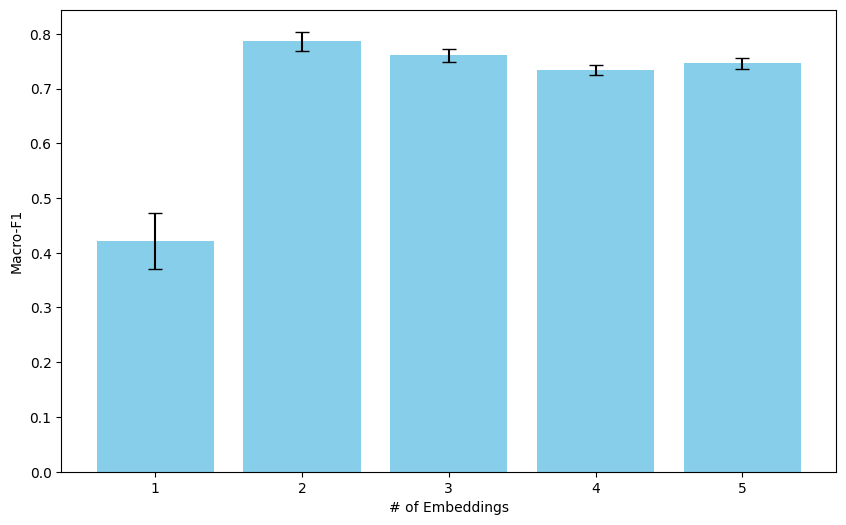}
        \subcaption{Factuality - ACL2020.}
    \end{minipage}
    
    \vspace{0.1cm}
    
    \begin{minipage}{0.54\linewidth}
    
        \centering
            \vspace{0.10cm}

        \includegraphics[width=\textwidth]{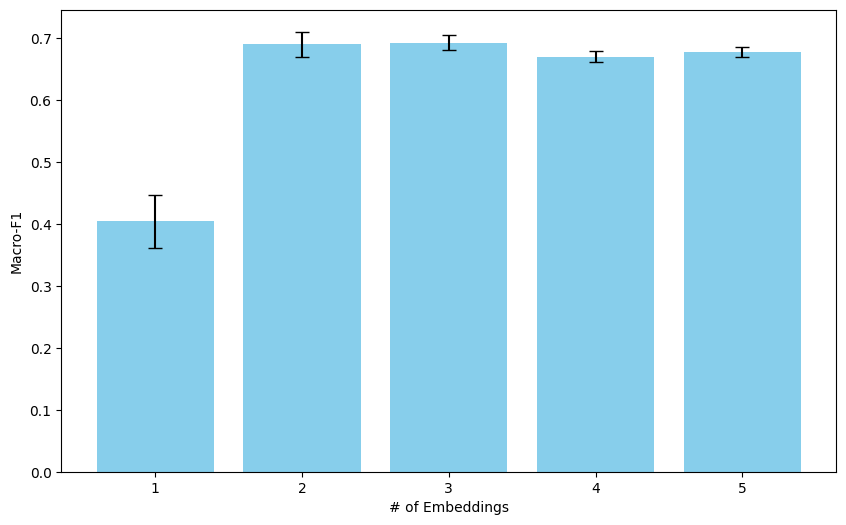}
        \subcaption{Political Bias - MBFC2024.}
    \end{minipage}
    \hfill
    \begin{minipage}{0.54\linewidth}
        \centering
        \includegraphics[width=\textwidth]{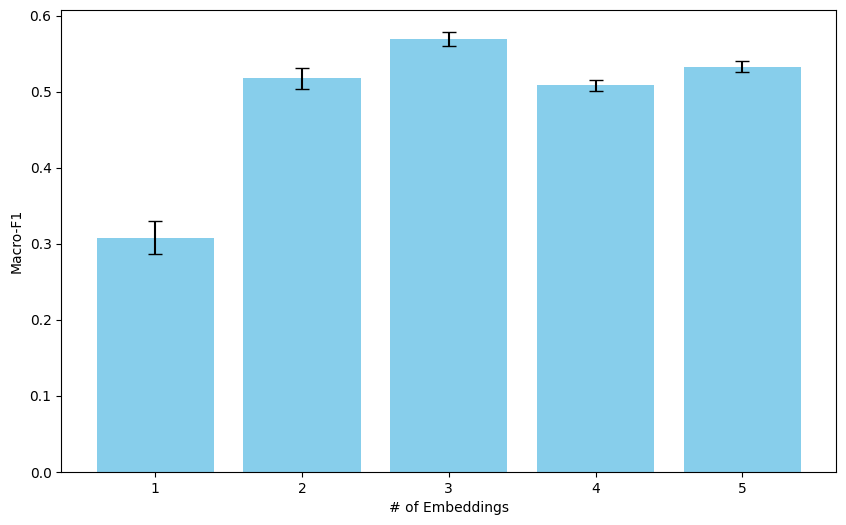}
        \subcaption{Factuality - MBFC2024.}
    \end{minipage}
    \caption{Dependency of \emph{Macro-F1} on the number of representations, with a $0.9$ confidence interval.}
     \label{barplot}
\end{figure*}

In Figure~\ref{barplot} we illustrate how the number of representations affects the results.
Overall, 2-3 representations are mainly enough; adding more often results in considerably lower gains.

\clearpage

\clearpage

\begin{table*}[!t]
\centering
\resizebox{1.0\textwidth}{!}{
\large 
\begin{tabular}{l|ccccc|ccccc}
\toprule
& \multicolumn{5}{c|}{\textbf{Political Bias}} & \multicolumn{5}{c}{\textbf{Factuality}} \\
\cmidrule{2-11}
\textbf{Models} & \textbf{MAE} & \textbf{Macro-F1} & \textbf{Accuracy} & \textbf{Precision} & \textbf{Recall} & \textbf{MAE} & \textbf{Macro-F1} & \textbf{Accuracy} & \textbf{Precision} & \textbf{Recall} \\
\midrule
 & & & & \textbf{ACL-2020} & & & & & \\
\midrule
Majority class & \textbf{0.633} & \textbf{17.91} & \textbf{36.73} & \textbf{12.24} & \textbf{33.33} & \textbf{0.571} & \textbf{24.79} & \textbf{59.18} & \textbf{19.73} & \textbf{33.33} \\
Middle class & 1.061 & 14.81 & 28.57 & 9.52 & \textbf{33.33} & 1.429 & 9.36 & 16.33 & 5.44 & \textbf{33.33} \\
\midrule
 & & & & \textbf{MBFC-2025}  & & & & & \\
\midrule
Majority class & \textbf{1.071} & \textbf{9.21} & \textbf{29.92} & \textbf{5.98} & \textbf{20.00} & \textbf{0.567} & \textbf{13.81} & \textbf{52.76} & \textbf{10.55} & \textbf{20.00} \\
Middle class & 1.449 & 5.68 & 16.54 & 3.31 & \textbf{20.00} & 1.504 & 2.09 & 5.51 & 1.10 & \textbf{20.00} \\

\bottomrule
\end{tabular}
}
\vspace{-0.5em}
\caption{Evaluation results for \emph{political bias} and \emph{factuality} using dummy classifiers for the majority and middle classes in the ACL-2020 and MBFC-2025 datasets.}
\label{baselines}
\end{table*}

\begin{table*}[t!]
\centering
\resizebox{1.0\textwidth}{!}{
\begin{tabular}{@{}llccccc@{}}
\toprule
\textbf{Dataset} & \textbf{Model} & \textbf{MAE} & \textbf{Macro-F1} & \textbf{Accuracy} & \textbf{Precision} & \textbf{Recall} \\ 
\midrule
ACL-2020 Bias & GPT-4o\textsubscript{0} & 0.562 & 62.50 & 54.37 & 60.83 & 49.33\\
              & GPT-4o\textsubscript{1} & 0.250 & 77.08 & 59.88 & 60.00 & 60.49\\
              & GPT-4o\textsubscript{3} & \textbf{0.229} & \textbf{79.17} & \textbf{61.40} & \textbf{60.86} & \textbf{62.28} \\
              & GPT-4o\textsubscript{5} & 0.250 & 77.08 & 59.88 & 60.00 & 60.49 \\
\midrule
ACL-2020 Factuality & GPT-4o\textsubscript{0} & 0.625 & 39.58 & 36.77 & 60.83 & 45.41 \\
                  & GPT-4o\textsubscript{1} & \textbf{0.766} & \textbf{42.86} & \textbf{35.69} & \textbf{33.02} & \textbf{43.43} \\
                  & GPT-4o\textsubscript{3} & 0.667 & 39.58 & 38.19 & 55.23 & 46.65 \\
                  & GPT-4o\textsubscript{5} & 0.729 & 29.17 & 24.37 & 47.44 & 32.28 \\
\midrule
MBFC-2025 Bias    & GPT-4o\textsubscript{0}& \textbf{0.766} & \textbf{42.86} & \textbf{35.69} & \textbf{33.02} & \textbf{43.43} \\
                   & GPT-4o\textsubscript{1} & 2.065 & 11.69 & 4.19 & 2.34 & 20.00 \\
                   & GPT-4o\textsubscript{3} & 2.065 & 11.69 & 4.19 & 2.34 & 20.00 \\
                   & GPT-4o\textsubscript{5} & 2.065 & 11.69 & 4.19 & 2.34 & 20.00 \\
\midrule
MBFC-2025 Factuality & GPT-4o\textsubscript{0} & 1.468 & 9.09 & 6.32 & 10.62 & 16.78\\
                     & GPT-4o\textsubscript{1} & \textbf{1.416} & \textbf{10.39} & \textbf{6.67} & 8.20 & 17.47 \\
                     & GPT-4o\textsubscript{3} & 2.169 & 3.90 & 4.62 & 4.66 & \textbf{27.36}\\
                     & GPT-4o\textsubscript{5} & 2.494 & 2.60 & 1.85 & \textbf{20.26} & 20.69 \\
\bottomrule
\end{tabular}
}
\caption{Results for \emph{political bias} and \emph{factuality} on ACL-2020 and MBFC-2025 media articles datasets using GPT-4o hard voting. The subscript denotes the number of few-shot examples used in the prompt.}
\label{LLM}
\end{table*}

\begin{table*}[!t]
\centering
\resizebox{1.0\textwidth}{!}{
\large 
\begin{tabular}{c|ccccc|ccccc}
\toprule
& \multicolumn{5}{c|}{\textbf{Hard Voting}} & \multicolumn{5}{c}{\textbf{Soft Voting}} \\
\cmidrule{2-11}
\textbf{Models} & \textbf{MAE} & \textbf{Macro-F1} & \textbf{Accuracy} & \textbf{Precision} & \textbf{Recall} & \textbf{MAE} & \textbf{Macro-F1} & \textbf{Accuracy} & \textbf{Precision} & \textbf{Recall} \\
\midrule
 & & & & & \textbf{ACL-2020} & & & & & \\ 
\midrule
 & & & & & \textbf{PLMs} & & & & & \\
\midrule
SVM$_{\mathrlap{\text{TF-IDF}}}$\hspace{16mm} & 0.571 & 37.00 & 44.89 & 63.49 & 42.54 & 0.551 & 33.86 & 44.89 & 46.66 & 42.85 \\
BERT$_{\mathrlap{\text{Base}}}$\hspace{14.8mm} & 0.346 & 69.03 & 69.38 & 72.44 & 68.39 & 0.367 & 68.18 & 69.38 & 72.62 & 67.86 \\
RoBERTa$_{\mathrlap{\text{Base}}}$\hspace{9.6mm}  & 0.306 & 75.93 & \textbf{77.55} & 75.91 & \textbf{75.93} & \textbf{0.285} & \textbf{76.16} & \textbf{77.55} & \textbf{80.51} & 75.91 \\
DistilBERT$_{\mathrlap{\text{Base}}}$\hspace{7.8mm} & 0.346 & 68.54 & 69.38 & 72.59 & 67.97 & 0.306 & 72.99 & 73.46 & 77.62 & 72.20 \\
ALBERT$_{\mathrlap{\text{Base}}}$\hspace{11mm}  & 0.510 & 62.12 & 65.30 & 64.39 & 63.10 & 0.428 & 67.05 & 69.38 & 70.41 & 67.44 \\
Ensemble\hspace{11mm} & 0.416 & 63.52 & 66.66 & 71.82 & 64.21 & 0.416 & 63.52 & 66.66 & 71.82 & 64.21 \\

\midrule
 & & & & & \textbf{LLMs} & & & & & \\
\midrule

LLaMA2 $_{\mathrlap{\text{0}}}$ $\hspace{9mm}$ & 1.933 & 11.05 & 11.94 & 11.15 & 13.13 & 1.930 & 11.59 & 24.17 & 16.44 & 24.17 \\
LLaMA2 $_{\mathrlap{\text{1}}}$ $\hspace{9mm}$ & 1.314 & 28.39 & 38.74 & 22.83 & 38.74 & 1.307 & 29.10 & 39.74 & 23.53 & 39.74 \\
LLaMA2 $_{\mathrlap{\text{3}}}$ $\hspace{9mm}$ & \textbf{1.201} & 34.42 & 41.72 & 34.88 & 41.72 & 1.182 & \textbf{36.41} & \textbf{44.37} & 35.85 & \textbf{44.37} \\
LLaMA2 $_{\mathrlap{\text{5}}}$ $\hspace{9mm}$ & 1.314 & 24.13 & 32.45 & 35.32 & 32.45 & 1.317 & 24.57 & 33.11 & 37.04 & 33.11 \\
Mistral $_{\mathrlap{\text{0}}}$ \hspace{11.8mm} & 1.923 & 9.53 & 23.51 & 7.47 & 23.51 & 1.894 & 12.49 & 25.17 & 23.86 & 25.17 \\
Mistral $_{\mathrlap{\text{1}}}$ \hspace{11.8mm} & 1.390 & 12.34 & 21.19 & 35.14 & 21.19 & 1.384 & 14.45 & 22.52 & 37.12 & 22.52 \\
Mistral $_{\mathrlap{\text{3}}}$ \hspace{11.8mm}  & 1.357 & 18.10 & 25.17 & 35.13 & 25.17 & 1.341 & 20.79 & 27.15 & \textbf{36.52} & 27.15 \\
Mistral $_{\mathrlap{\text{5}}}$ \hspace{11.8mm} & 1.417 & 7.01 & 18.21 & 22.87 & 18.21 & 1.417 & 7.01 & 18.21 & 22.87 & 18.21 \\
Ensemble $_{\mathrlap{\text{0}}}$ \hspace{8.3mm}  & 1.930 & 9.58 & 23.51 & 15.47 & 23.51 & 1.943 & 9.44 & 23.51 & 15.36 & 23.51 \\
Ensemble $_{\mathrlap{\text{1}}}$ \hspace{8.3mm}  & 1.407 & 22.43 & 28.48 & 22.69 & 28.48 & 1.205 & 35.19 & 44.37 & 34.46 & 44.37 \\
Ensemble $_{\mathrlap{\text{3}}}$ \hspace{8.3mm}  & 1.278 & 27.35 & 32.78 & 36.28 & 32.78 & 1.238 & 31.79 & 38.08 & 37.14 & 38.08 \\
Ensemble $_{\mathrlap{\text{5}}}$ \hspace{8.3mm} & 1.387 & 15.78 & 23.84 & 33.12 & 23.84 & 1.324 & 21.85 & 28.81 & 36.05 & 28.81 \\

\midrule
 & & & & & \textbf{MBFC-2025} & & & & & \\
\midrule
 & & & & & \textbf{PLMs} & & & & & \\
 
\midrule
SVM$_{\mathrlap{\text{TF-IDF}}}$\hspace{16mm} & 0.630 & 50.80 & 58.66 & 60.65 & 50.20 & 0.591 & 47.74 & 57.87 & 65.65 & 47.28 \\
BERT$_{\mathrlap{\text{Base}}}$\hspace{14.8mm} & 0.539 & 63.85 & 68.50 & 65.55 & 63.11 & 0.571 & 61.51 & 65.75 & 62.98 & 60.66 \\
RoBERTa$_{\mathrlap{\text{Base}}}$\hspace{9.6mm} & 0.484 & 66.33 & 70.47 & 66.81 & 66.38 & 0.528 & 64.51 & 69.69 & 65.58 & 64.53 \\
DistilBERT$_{\mathrlap{\text{Base}}}$\hspace{7.8mm} & 0.547 & 63.95 & 68.90 & 66.74 & 63.35 & 0.583 & 62.27 & 67.32 & 64.45 & 61.59 \\
ALBERT$_{\mathrlap{\text{Base}}}$\hspace{11mm} & 0.484 & 68.93 & 72.05 & 69.68 & 69.00 & \textbf{0.457} & \textbf{71.74} & \textbf{74.41} & \textbf{71.84} & \textbf{71.61} \\
Ensemble\hspace{11mm} & 0.465 & 69.48 & 72.44 & 70.37 & 69.20 & 0.472 & 68.71 & 72.05 & 69.48 & 68.43 \\

\midrule

 & & & & & \textbf{LLMs} & & & & & \\
\midrule

LLaMA2 $_{\mathrlap{\text{0}}}$ $\hspace{9mm}$ & 1.933 & 11.05 & 23.84 & 15.33 & 23.84 & 1.930 & 11.59 & 24.17 & \textbf{16.44} & 24.17 \\
LLaMA2 $_{\mathrlap{\text{1}}}$ $\hspace{9mm}$ & \textbf{1.130} & 14.38 & 19.51 & 14.28 & 19.51 & 1.122 & 14.52 & 20.33 & 14.45 & 20.33 \\
LLaMA2 $_{\mathrlap{\text{3}}}$ $\hspace{9mm}$ & 1.789 & 12.27 & 21.14 & 10.61 & 21.14 & 1.821 & 11.87 & 20.33 & 11.50 & 20.33 \\
LLaMA2 $_{\mathrlap{\text{5}}}$ $\hspace{9mm}$ & 1.122 & \textbf{17.17} & 22.76 & 15.84 & 22.76 & 1.130 & 17.13 & 22.76 & 15.63 & 22.76 \\
Mistral $_{\mathrlap{\text{0}}}$ \hspace{11.8mm} & 1.924 & 9.53 & 23.51 & 7.47 & 23.51 & 1.894 & 12.49 & 25.17 & 23.86 & 25.17 \\
Mistral $_{\mathrlap{\text{1}}}$ \hspace{11.8mm} & 1.512 & 9.71 & 21.95 & 10.81 & 21.95 & 1.504 & 10.82 & 22.76 & 11.66 & 22.76 \\
Mistral $_{\mathrlap{\text{3}}}$ \hspace{11.8mm}  & 1.480 & 13.50 & 23.58 & 14.35 & 23.58 & 1.480 & 14.61 & 24.39 & 16.20 & 24.39 \\
Mistral $_{\mathrlap{\text{5}}}$ \hspace{11.8mm} & 1.447 & 11.54 & 23.58 & 11.61 & 23.58 & 1.439 & 12.33 & 24.39 & 12.27 & 24.39 \\
Ensemble $_{\mathrlap{\text{0}}}$ \hspace{8.3mm}  & 1.930 & 9.58 & 23.51 & 15.47 & 23.51 & 1.944 & 9.44 & 23.51 & 15.36 & 23.51 \\
Ensemble $_{\mathrlap{\text{1}}}$ \hspace{8.3mm}  & 1.325 & 15.41 & 24.39 & 15.23 & 24.39 & 1.333 & 15.25 & \textbf{25.20} & 12.62 & \textbf{25.20} \\
Ensemble $_{\mathrlap{\text{3}}}$ \hspace{8.3mm}  & 1.602 & 13.27 & 22.76 & 9.96 & 22.76 & 1.683 & 14.11 & 24.39 & 10.67 & 24.39 \\
Ensemble $_{\mathrlap{\text{5}}}$ \hspace{8.3mm} & 1.325 & 12.93 & 21.95 & 12.33 & 21.95 & 1.211 & 14.92 & 21.95 & 13.76 & 21.95 \\

\bottomrule

\end{tabular}
}
\vspace{-0.5em}
\caption{Evaluation results for \emph{political bias} using hard-  and soft-votings for each framework and ensemble in ensemble.}
\label{Task 1A}
\end{table*}

\begin{table*}[!t]
\centering
\resizebox{1.0\textwidth}{!}{
\large 
\begin{tabular}{c|ccccc|ccccc}
\toprule
& \multicolumn{5}{c|}{\textbf{Hard Voting}} & \multicolumn{5}{c}{\textbf{Soft Voting}} \\
\cmidrule{2-11}
\textbf{Models} & \textbf{MAE} & \textbf{Macro-F1} & \textbf{Accuracy} & \textbf{Precision} & \textbf{Recall} & \textbf{MAE} & \textbf{Macro-F1} & \textbf{Accuracy} & \textbf{Precision} & \textbf{Recall} \\
\midrule
 & & & & & \textbf{ACL-2020} & & & & & \\ 
\midrule
 & & & & & \textbf{PLMs} & & & & & \\
\midrule
SVM$_{\mathrlap{\text{TF-IDF}}}$\hspace{16mm} & \textbf{0.449} & 48.43 & 65.31 & 72.81 & 47.94 & 0.551 & 30.24 & 61.22 & 53.47 & 36.11 \\
BERT$_{\mathrlap{\text{Base}}}$\hspace{14.8mm} & 0.510 & 32.37 & 59.18 & 30.95 & 36.59 & 0.551 & 32.50 & 59.18 & 33.79 & 36.59 \\
RoBERTa$_{\mathrlap{\text{Base}}}$\hspace{9.6mm} & 0.510 & 34.64 & 59.18 & 32.78 & 38.22 & 0.510 & 34.64 & 59.18 & 32.78 & 38.22 \\
DistilBERT$_{\mathrlap{\text{Base}}}$\hspace{7.8mm} & \textbf{0.449} & 34.26 & 63.27 & 33.59 & 38.89 & 0.510 & 30.22 & 61.22 & 32.13 & 36.11 \\
ALBERT$_{\mathrlap{\text{Base}}}$\hspace{11mm} & \textbf{0.449} & \textbf{51.46} & \textbf{67.35} & \textbf{76.14} & \textbf{50.72} & 0.469 & 47.16 & 65.31 & 74.33 & 46.31 \\
Ensemble\hspace{11mm} & 0.479 & 34.13 & 62.50 & 33.33 & 38.24 & 0.479 & 35.37 & 64.58 & 38.64 & 39.39 \\

\midrule
 & & & & & \textbf{LLMs} & & & & & \\
\midrule

LLaMA2 $_{\mathrlap{\text{0}}}$ $\hspace{9mm}$ & 2.095 & 14.60 & 23.84 & 13.74 & 23.84 & 2.263 & 10.17 & 19.19 & 12.71 & 19.19 \\
LLaMA2 $_{\mathrlap{\text{1}}}$ $\hspace{9mm}$ & 2.090 & \textbf{14.65} & \textbf{23.89} & 13.44 & \textbf{23.89} & 2.222 & 11.11 & 19.21 & 12.92 & 19.21 \\
LLaMA2 $_{\mathrlap{\text{3}}}$ $\hspace{9mm}$ & 2.393 & 5.36 & 16.09 & 9.97 & 16.09 & 2.386 & 5.93 & 16.41 & 13.74 & 16.41 \\
LLaMA2 $_{\mathrlap{\text{5}}}$ $\hspace{9mm}$ & 2.417 & 4.15 & 15.48 & 2.39 & 15.48 & 2.417 & 4.15 & 15.48 & 2.39 & 15.48 \\
Mistral $_{\mathrlap{\text{0}}}$ \hspace{11.8mm} & 2.120 & 12.55 & 22.12 & 12.19 & 22.12 & 2.221 & 8.99 & 18.33 & 10.74 & 18.33 \\
Mistral $_{\mathrlap{\text{1}}}$ \hspace{11.8mm} & 2.114 & 12.78 & 22.29 & 12.42 & 22.29 & 2.269 & 8.57 & 18.26 & 10.42 & 18.26 \\
Mistral $_{\mathrlap{\text{3}}}$ \hspace{11.8mm}  & 2.374 & 6.45 & 16.71 & 12.50 & 16.71 & 2.380 & 6.48 & 16.71 & \textbf{15.36} & 16.71 \\
Mistral $_{\mathrlap{\text{5}}}$ \hspace{11.8mm} & 1.739 & 6.29 & 18.58 & 5.98 & 18.58 & \textbf{1.733} & 6.28 & 18.58 & 7.25 & 18.58 \\
Ensemble $_{\mathrlap{\text{0}}}$ \hspace{8.3mm}  & 2.111 & 14.00 & 23.21 & 13.33 & 23.21 & 2.277 & 10.00 & 19.31 & 13.78 & 19.31 \\
Ensemble $_{\mathrlap{\text{1}}}$ \hspace{8.3mm}  & 2.102 & 14.25 & 23.52 & 13.74 & 23.52 & 2.269 & 10.39 & 19.50 & 13.91 & 19.50 \\
Ensemble $_{\mathrlap{\text{3}}}$ \hspace{8.3mm}  & 2.374 & 6.45 & 16.71 & 12.50 & 16.71 & 2.405 & 4.77 & 15.79 & 9.95 & 15.79 \\
Ensemble $_{\mathrlap{\text{5}}}$ \hspace{8.3mm} & 2.080 & 7.65 & 15.17 & 5.13 & 15.17 & 2.402 & 4.21 & 15.48 & 2.43 & 15.48 \\

\midrule
 & & & & & \textbf{MBFC-2025} & & & & & \\
\midrule

 & & & & & \textbf{PLMs} & & & & & \\
\midrule
SVM$_{\mathrlap{\text{TF-IDF}}}$\hspace{16mm} & 0.354 & 29.90 & 68.90 & 33.92 & 30.07 & 0.433 & 23.99 & 64.17 & 29.24 & 25.98 \\
BERT$_{\mathrlap{\text{Base}}}$\hspace{14.8mm} & 0.276 & 37.28 & 75.59 & 40.17 & 36.76 & 0.287 & 37.62 & 75.20 & \textbf{44.75} & 36.56 \\
RoBERTa$_{\mathrlap{\text{Base}}}$\hspace{9.6mm} & 0.260 & \textbf{41.12} & 75.98 & 42.71 & \textbf{40.31} & 0.280 & 34.60 & 74.41 & 35.21 & 34.64 \\
DistilBERT$_{\mathrlap{\text{Base}}}$\hspace{7.8mm} & \textbf{0.252} & 38.03 & \textbf{77.56} & 43.23 & 37.00 & 0.276 & 35.61 & 76.38 & 40.72 & 35.10 \\
ALBERT$_{\mathrlap{\text{Base}}}$\hspace{11mm} & \textbf{0.252} & 35.57 & 76.38 & 37.30 & 35.50 & 0.264 & 35.50 & 75.98 & 38.28 & 35.29 \\
Ensemble\hspace{11mm} & 0.256 & 37.95 & 77.17 & 42.62 & 37.08 & 0.260 & 39.73 & 76.77 & 46.32 & 38.15 \\

\midrule
 & & & & & \textbf{LLMs} & & & & & \\
\midrule

LLaMA2 $_{\mathrlap{\text{0}}}$ $\hspace{9mm}$ & 1.541 & 14.57 & 23.84 & 13.95 & 23.84 & \textbf{1.535} & \textbf{15.25} & \textbf{24.77} & 14.45 & \textbf{24.77} \\
LLaMA2 $_{\mathrlap{\text{1}}}$ $\hspace{9mm}$ & 2.516 & 2.97 & 12.90 & 1.67 & 12.90 & 2.516 & 2.95 & 12.90 & 1.66 & 12.90 \\
LLaMA2 $_{\mathrlap{\text{3}}}$ $\hspace{9mm}$ & 2.516 & 2.95 & 12.90 & 1.66 & 12.90 & 2.516 & 2.95 & 12.90 & 1.66 & 12.90 \\
LLaMA2 $_{\mathrlap{\text{5}}}$ $\hspace{9mm}$ & 2.516 & 2.95 & 12.90 & 1.66 & 12.90 & 2.516 & 2.95 & 12.90 & 1.66 & 12.90 \\
Mistral $_{\mathrlap{\text{0}}}$ \hspace{11.8mm} & 1.761 & 6.06 & 17.65 & 4.51 & 17.65 & 1.736 & 6.20 & 18.26 & 5.93 & 18.26 \\
Mistral $_{\mathrlap{\text{1}}}$ \hspace{11.8mm} & 2.419 & 4.15 & 13.71 & 3.26 & 13.71 & 2.483 & 3.01 & 12.90 & 1.71 & 12.90 \\
Mistral $_{\mathrlap{\text{3}}}$ \hspace{11.8mm}  & 2.516 & 2.97 & 12.90 & 1.67 & 12.90 & 2.516 & 2.95 & 12.90 & 1.66 & 12.90 \\
Mistral $_{\mathrlap{\text{5}}}$ \hspace{11.8mm} & 1.782 & 1.32 & 8.06 & 0.72 & 8.06 & 1.790 & 1.33 & 8.06 & 0.73 & 8.06 \\
Ensemble $_{\mathrlap{\text{0}}}$ \hspace{8.3mm}  & 1.705 & 12.65 & 17.03 & \textbf{15.20} & 17.03 & 1.770 & 11.99 & 18.89 & 13.68 & 18.89 \\
Ensemble $_{\mathrlap{\text{1}}}$ \hspace{8.3mm}  & 2.467 & 3.03 & 12.90 & 1.72 & 12.90 & 2.516 & 2.95 & 12.90 & 1.66 & 12.90 \\
Ensemble $_{\mathrlap{\text{3}}}$ \hspace{8.3mm}  & 2.516 & 2.95 & 12.90 & 1.66 & 12.90 & 2.516 & 2.95 & 12.90 & 1.66 & 12.90 \\
Ensemble $_{\mathrlap{\text{5}}}$ \hspace{8.3mm} & 2.112 & 3.86 & 10.48 & 2.40 & 10.48 & 2.459 & 3.10 & 12.90 & 1.76 & 12.90 \\

\bottomrule
\end{tabular}
}
\vspace{-0.5em}
\caption{Evaluation results for \emph{factuality} using hard and soft voting for each framework and in ensemble.}
\label{Task 1B}
\end{table*}

\begin{table*}[!t]
\centering
\resizebox{1.0\textwidth}{!}{
\large 
\begin{tabular}{l|ccccc|ccccc}
\toprule
& \multicolumn{5}{c|}{\textbf{Political Bias}} & \multicolumn{5}{c}{\textbf{Factuality}} \\
\cmidrule{2-11}
\textbf{Models} & \textbf{MAE} & \textbf{Macro-F1} & \textbf{Accuracy} & \textbf{Precision} & \textbf{Recall} & \textbf{MAE} & \textbf{Macro-F1} & \textbf{Accuracy} & \textbf{Precision} & \textbf{Recall} \\

\midrule
 & & & & & \textbf{ACL-2020} & & &  & & \\
\midrule
 & & & & & \textbf{PLMs} & & &  & & \\
\midrule
SVM$_{\mathrlap{\text{TF-IDF}}}$ & 0.633 & 52.23 & 53.06 & 52.38 & 52.61 & 0.551 & \textbf{36.92} & 57.14 & 34.12 & \textbf{40.33} \\
BERT$_{\mathrlap{\text{Base}}}$ & 0.673 & 41.62 & 48.98 & 45.66 & 47.31 & \textbf{0.531} & 30.20 & \textbf{61.22} & 37.23 & 36.11 \\
RoBERTa$_{\mathrlap{\text{Base}}}$  & \textbf{0.571} & \textbf{57.52} & \textbf{59.18} & \textbf{58.14} & \textbf{57.86} & 0.571 & 24.79 & 59.18 & 19.73 & 33.33 \\
DistilBERT$_{\mathrlap{\text{Base}}}$ & 0.653 & 24.63 & 38.78 & 29.55 & 35.40 & 0.571 & 24.79 & 59.18 & 19.73 & 33.33 \\
ALBERT$_{\mathrlap{\text{Base}}}$ & 0.592 & 47.68 & 53.06 & 54.67 & 50.81 & 0.571 & 28.49 & 57.14 & 28.33 & 33.81 \\
Ensemble HV & 0.604 & 44.63 & 50.00 & 54.58 & 48.74 & 0.542 & 29.99 & 60.42 & \textbf{36.96} & 36.11 \\
Ensemble SV & 0.583 & 55.00 & 56.25 & 56.62 & 55.46 & 0.583 & 24.56 & 58.33 & 19.44 & 33.33 \\
\midrule
 & & & & & \textbf{LLMs} & & & & & \\
\midrule
LLaMA2 & \textbf{0.568} & 36.64 & \textbf{45.95} & 65.14 & 45.95 & \textbf{1.536} & \textbf{15.25} & 12.68 & 14.45 & \textbf{24.77} \\
Mistral & 0.746 & 40.65 & 45.74 & 61.17 & 45.74 & 1.737 & 6.07 & \textbf{27.86} & 5.93 & 18.27 \\
Ensemble HV & 0.640 & \textbf{42.73} & 48.02 & \textbf{62.22} & \textbf{48.02} & 1.706 & 12.65 & 21.62 & \textbf{15.20} & 17.03 \\

Ensemble SV & 0.655 & 41.11 & 46.19 & 59.86 & 46.19 & 1.706 & 12.17 & 20.80 & 14.63 & 16.39 \\

\midrule
 & & & & & \textbf{MBFC-2025} & & & & & \\
\midrule
 & & & & & \textbf{PLMs} & & & & & \\
\midrule
SVM$_{\mathrlap{\text{TF-IDF}}}$ & 0.771 & 52.82 & 56.07 & 55.58 & 52.41 & 0.552 & 25.23 & 50.25 & \textbf{27.02} & 25.09 \\
BERT$_{\mathrlap{\text{Base}}}$ & 0.645 & 60.86 & 62.15 & 63.86 & 60.42 & \textbf{0.398} & 27.04 & 63.68 & 25.49 & 28.80 \\
RoBERTa$_{\mathrlap{\text{Base}}}$ & 0.692 & 58.98 & 60.75 & 59.29 & 59.28 & 0.413 & \textbf{29.49} & 63.68 & 45.47 & \textbf{30.04} \\
DistilBERT$_{\mathrlap{\text{Base}}}$ & 0.766 & 54.14 & 58.88 & 61.82 & 54.01 & \textbf{0.398} & 27.45 & \textbf{64.68} & 26.07 & 29.18 \\
ALBERT$_{\mathrlap{\text{Base}}}$ & 0.734 & 56.42 & 59.35 & 59.11 & 57.63 & 0.413 & 27.22 & 64.18 & 25.67 & 29.03 \\
Ensemble HV & \textbf{0.623} & \textbf{63.19} & \textbf{64.15} & \textbf{65.33} & \textbf{62.86} & 0.403 & 27.26 & 64.18 & 25.74 & 28.99 \\
Ensemble SV & 0.698 & 58.30 & 60.85 & 60.74 & 58.04 & 0.403 & 27.01 & 63.68 & 25.45 & 28.82 \\
\midrule
 & & & & & \textbf{LLMs} & & & & & \\
\midrule
LLaMA2 & 1.424 & 18.95 & 31.39 & \textbf{27.31} & 31.39 & 1.011 & 32.79 & 40.95 & 27.69 & 40.95 \\
Mistral & 1.418 & 9.06 & 14.86 & 21.88 & 14.86 & \textbf{0.849} & \textbf{36.06} & \textbf{51.21} & 28.12 & \textbf{51.21} \\
Ensemble HV & \textbf{1.411} & 16.01 & 24.15 & 24.30 & 24.15 & 0.927 & 34.72 & 46.21 & 27.83 & 46.21 \\
Ensemble SV  & 1.426 & \textbf{20.19} & \textbf{32.13} & 27.07 & \textbf{32.13} & 0.872 & 36.01 & 49.24 & \textbf{28.39} & 49.24 \\
\bottomrule
\end{tabular}
}
\vspace{-0.5em}
\caption{Evaluation results for \emph{political bias} and \emph{factuality} using frameworks independently and in ensembles using hard voting (HV) and soft voting (SV) at media-level Wikipedia datasets.}
\label{Task 2}
\end{table*}


\begin{table*}[!t]
\centering
\resizebox{1.0\textwidth}{!}{
\large 
\begin{tabular}{c|ccccc|ccccc}
\toprule
& \multicolumn{5}{c|}{\textbf{Hard Voting}} & \multicolumn{5}{c}{\textbf{Soft Voting}} \\
\cmidrule{2-11}
\textbf{Models} & \textbf{MAE} & \textbf{Macro-F1} & \textbf{Accuracy} & \textbf{Precision} & \textbf{Recall} & \textbf{MAE} & \textbf{Macro-F1} & \textbf{Accuracy} & \textbf{Precision} & \textbf{Recall} \\

\midrule
 & & & & & \textbf{ACL-2020} & & & & & \\ 
\midrule

 & & & & & \textbf{PLMs} & & & & & \\
\midrule

SVM$_{\mathrlap{\text{TF-IDF}}}$\hspace{16mm} & 0.490 & 47.36 & 53.06 & 76.11 & 51.12 & 0.490 & 50.00 & 55.10 & 69.36 & 53.50 \\
BERT$_{\mathrlap{\text{Base}}}$\hspace{14.8mm} & 0.367 & 68.69 & 69.39 & 72.82 & 68.29 & 0.367 & 66.84 & 67.35 & 72.99 & 66.32 \\
RoBERTa$_{\mathrlap{\text{Base}}}$\hspace{9.6mm} & \textbf{0.224} & \textbf{81.47} & \textbf{81.63} & \textbf{83.61} & \textbf{81.20} & 0.286 & 77.16 & 77.55 & 78.45 & 76.97 \\
DistilBERT$_{\mathrlap{\text{Base}}}$\hspace{7.8mm} & 0.327 & 70.45 & 71.43 & 76.35 & 69.83 & 0.388 & 65.67 & 67.35 & 72.77 & 65.48 \\
ALBERT$_{\mathrlap{\text{Base}}}$\hspace{11mm} & 0.551 & 56.63 & 61.22 & 60.04 & 58.65 & 0.449 & 62.02 & 65.31 & 67.86 & 62.99 \\
Ensemble\hspace{11mm} & 0.375 & 67.76 & 68.75 & 71.20 & 67.36 & 0.354 & 70.02 & 70.83 & 73.64 & 69.51 \\

\midrule
 & & & & & \textbf{LLMs} & & & & & \\
\midrule

LLaMA2 $_{\mathrlap{\text{0}}}$ $\hspace{9mm}$ & 1.500 & 16.64 & 20.00 & 32.30 & 20.00 & 1.510 & 15.54 & 19.67 & 32.16 & 19.67 \\
LLaMA2 $_{\mathrlap{\text{1}}}$ $\hspace{9mm}$ & 1.673 & 19.53 & 32.00 & 15.01 & 32.00 & 1.657 & 19.22 & 34.00 & 16.39 & 34.00 \\
LLaMA2 $_{\mathrlap{\text{3}}}$ $\hspace{9mm}$ & \textbf{1.233} & 20.08 & 25.00 & 23.47 & 25.00 & 1.237 & 22.13 & 28.00 & 24.69 & 28.00 \\
LLaMA2 $_{\mathrlap{\text{5}}}$ $\hspace{9mm}$ & 1.559 & 9.75 & 14.63 & 21.18 & 14.63 & 1.537 & 11.60 & 15.85 & 24.48 & 15.85 \\
Mistral $_{\mathrlap{\text{0}}}$ \hspace{11.8mm}  & 1.440 & 10.73 & 24.67 & 19.74 & 24.67 & 1.400 & 12.77 & 24.33 & 19.53 & 24.33 \\
Mistral $_{\mathrlap{\text{1}}}$ \hspace{11.8mm} & 1.453 & 20.22 & 22.00 & 33.03 & 22.00 & 1.457 & 20.92 & 23.00 & 32.05 & 23.00 \\
Mistral $_{\mathrlap{\text{3}}}$ \hspace{11.8mm}  & 1.493 & 19.66 & 21.00 & 36.86 & 21.00 & 1.483 & 18.35 & 20.00 & \textbf{36.18} & 20.00 \\
Mistral $_{\mathrlap{\text{5}}}$ \hspace{11.8mm} & 1.375 & 21.43 & 23.17 & 36.21 & 23.17 & 1.358 & 20.07 & 21.95 & 36.31 & 21.95 \\
Ensemble $_{\mathrlap{\text{0}}}$ \hspace{8.3mm}  & 1.460 & 16.61 & 24.00 & 25.24 & 24.00 & 1.490 & 17.51 & 21.67 & 31.73 & 21.67 \\
Ensemble $_{\mathrlap{\text{1}}}$ \hspace{8.3mm}  & 1.590 & 21.38 & 27.00 & 20.62 & 27.00 & 1.350 & \textbf{28.87} & \textbf{35.00} & 26.08 & \textbf{35.00} \\
Ensemble $_{\mathrlap{\text{3}}}$ \hspace{8.3mm}  & 1.363 & 20.95 & 23.00 & 27.97 & 23.00 & 1.307 & 21.86 & 25.00 & 26.61 & 25.00 \\
Ensemble $_{\mathrlap{\text{5}}}$ \hspace{8.3mm} & 1.452 & 17.21 & 19.51 & 37.04 & 19.51 & 1.537 & 11.60 & 15.85 & 24.48 & 15.85 \\

\midrule
 & & & & & \textbf{MBFC-2025} & & & & & \\

\midrule
 & & & & & \textbf{PLMs} & & & & & \\
\midrule
SVM$_{\mathrlap{\text{TF-IDF}}}$\hspace{16mm} & 0.712 & 48.47 & 58.02 & 64.04 & 49.03 & 0.708 & 47.63 & 57.55 & 63.41 & 48.28 \\
BERT$_{\mathrlap{\text{Base}}}$\hspace{14.8mm} & 0.566 & 66.21 & 69.81 & 69.04 & 66.02 & 0.491 & 69.19 & 72.17 & 70.38 & 68.70 \\
RoBERTa$_{\mathrlap{\text{Base}}}$\hspace{9.6mm} & 0.500 & 66.54 & 70.28 & 67.68 & 66.43 & 0.505 & 67.03 & 70.75 & 68.19 & 66.75 \\
DistilBERT$_{\mathrlap{\text{Base}}}$\hspace{7.8mm} & 0.519 & \textbf{67.13} & \textbf{71.23} & \textbf{68.81} & \textbf{67.43} & 0.495 & 66.03 & 70.75 & 67.99 & 66.26 \\
ALBERT$_{\mathrlap{\text{Base}}}$\hspace{11mm} & 0.500 & 68.23 & 71.23 & 69.59 & 68.60 & 0.538 & 68.32 & 71.70 & 69.76 & 68.99 \\
Ensemble\hspace{11mm} & \textbf{0.448} & 70.23 & 73.58 & 72.09 & 70.37 & \textbf{0.425} & \textbf{72.41} & \textbf{75.47} & \textbf{73.69} & \textbf{72.36} \\

\midrule
 & & & & & \textbf{LLMs} & & & & & \\
\midrule

LLaMA2 $_{\mathrlap{\text{0}}}$ $\hspace{9mm}$ & 1.500 & 16.64 & 19.00 & \textbf{40.47} & 19.00 & 1.510 & 15.54 & 18.00 & 40.15 & 18.00 \\
LLaMA2 $_{\mathrlap{\text{1}}}$ $\hspace{9mm}$ & 1.673 & 18.22 & 29.67 & 14.71 & 29.67 & 1.657 & 19.22 & 30.67 & 15.61 & 30.67 \\
LLaMA2 $_{\mathrlap{\text{3}}}$ $\hspace{9mm}$ & \textbf{1.233} & \textbf{24.18} & \textbf{33.00} & 25.19 & \textbf{33.00} & 1.237 & 24.12 & 33.00 & 24.76 & 33.00 \\
LLaMA2 $_{\mathrlap{\text{5}}}$ $\hspace{9mm}$ & 1.559 & 4.23 & 12.37 & 7.32 & 12.37 & 1.549 & 5.34 & 13.38 & 8.85 & 13.38 \\
Mistral $_{\mathrlap{\text{0}}}$ \hspace{11.8mm}  & 1.440 & 10.73 & 22.00 & 7.18 & 22.00 & 1.400 & 10.39 & 22.00 & 6.82 & 22.00 \\
Mistral $_{\mathrlap{\text{1}}}$ \hspace{11.8mm} & 1.453 & 19.23 & 23.67 & 29.66 & 23.67 & 1.457 & 19.81 & 24.33 & 29.74 & 24.33 \\
Mistral $_{\mathrlap{\text{3}}}$ \hspace{11.8mm}  & 1.493 & 15.89 & 20.33 & \textbf{31.69} & 20.33 & 1.483 & 16.79 & 21.00 & 33.03 & 21.00 \\
Mistral $_{\mathrlap{\text{5}}}$ \hspace{11.8mm} & \textbf{1.375} & 19.93 & 25.75 & 26.94 & 25.75 & \textbf{1.358} & 20.32 & 26.76 & 26.80 & 26.76 \\
Ensemble $_{\mathrlap{\text{0}}}$ \hspace{8.3mm}  & 1.460 & 15.18 & 19.00 & 24.43 & 19.00 & 1.500 & 18.35 & 20.00 & 36.18 & 20.00 \\
Ensemble $_{\mathrlap{\text{1}}}$ \hspace{8.3mm}  & 1.520 & 20.59 & 28.67 & 17.69 & 28.67 & \textbf{1.320} & \textbf{26.61} & \textbf{36.67} & 22.07 & \textbf{36.67} \\
Ensemble $_{\mathrlap{\text{3}}}$ \hspace{8.3mm}  & 1.363 & 22.87 & 29.00 & 26.56 & 29.00 & 1.307 & 24.57 & 32.33 & 28.14 & 32.33 \\
Ensemble $_{\mathrlap{\text{5}}}$ \hspace{8.3mm} & 1.452 & 15.26 & 20.74 & 27.40 & 20.74 & 1.498 & 10.71 & 18.06 & 11.41 & 18.06 \\

\bottomrule

\end{tabular}
}
\vspace{-0.5em}
\caption{Evaluation results for \emph{political bias} using hard and soft voting for each framework and in the ensemble.}
\label{Task 3A}
\end{table*}

\begin{table*}[!t]
\centering
\resizebox{1.0\textwidth}{!}{
\large 
\begin{tabular}{c|ccccc|ccccc}
\toprule
& \multicolumn{5}{c|}{\textbf{Hard Voting}} & \multicolumn{5}{c}{\textbf{Soft Voting}} \\
\cmidrule{2-11}
\textbf{Models} & \textbf{MAE} & \textbf{Macro-F1} & \textbf{Accuracy} & \textbf{Precision} & \textbf{Recall} & \textbf{MAE} & \textbf{Macro-F1} & \textbf{Accuracy} & \textbf{Precision} & \textbf{Recall} \\

\midrule
 & & & & & \textbf{ACL-2020} & & & & & \\ 
\midrule

 & & & & & \textbf{PLMs} & & & & & \\
\midrule
SVM$_{\mathrlap{\text{TF-IDF}}}$\hspace{16mm} & 0.571 & 24.79 & 59.18 & 19.73 & 33.33 & 0.571 & 24.79 & 59.18 & 19.73 & 33.33 \\
BERT$_{\mathrlap{\text{Base}}}$\hspace{14.8mm} & 0.531 & 25.44 & 59.18 & 20.57 & 33.33 & 0.551 & 25.11 & 59.18 & 20.14 & 33.33 \\
RoBERTa$_{\mathrlap{\text{Base}}}$\hspace{9.6mm} & 0.510 & 30.22 & 61.22 & 32.13 & 36.11 & 0.510 & 30.22 & 61.22 & 32.13 & 36.11 \\
DistilBERT$_{\mathrlap{\text{Base}}}$\hspace{7.8mm} & 0.551 & 30.24 & 61.22 & 53.47 & 36.11 & 0.551 & 30.24 & 61.22 & 53.47 & 36.11 \\
ALBERT$_{\mathrlap{\text{Base}}}$\hspace{11mm} & \textbf{0.469} & \textbf{42.42} & \textbf{65.31} & \textbf{77.04} & \textbf{43.06} & 0.510 & 34.67 & 63.27 & 43.24 & 38.89 \\
Ensemble\hspace{10.3mm} & 0.500 & 30.89 & 62.50 & 32.59 & 36.36 & 0.521 & 30.91 & 62.50 & 37.68 & 36.36 \\

\midrule
 & & & & & \textbf{LLMs} & & & & & \\
\midrule

LLaMA2 $_{\mathrlap{\text{0}}}$ $\hspace{9mm}$ & 1.990 & 10.52 & 20.00 & 7.83 & 20.00 & 1.950 & 11.59 & 22.00 & 8.46 & 22.00 \\
LLaMA2 $_{\mathrlap{\text{1}}}$ $\hspace{9mm}$ & 2.470 & 3.74 & 11.00 & 21.01 & 11.00 & 2.490 & 1.82 & 10.00 & 1.00 & 10.00 \\
LLaMA2 $_{\mathrlap{\text{3}}}$ $\hspace{9mm}$ & 1.430 & 20.08 & 25.00 & 23.47 & 25.00 & 1.390 & \textbf{22.13} & \textbf{28.00} & 24.69 & \textbf{28.00} \\
LLaMA2 $_{\mathrlap{\text{5}}}$ $\hspace{9mm}$ & 2.484 & 1.87 & 10.10 & 1.03 & 10.10 & 2.485 & 1.87 & 10.10 & 1.03 & 10.10 \\
Mistral $_{\mathrlap{\text{0}}}$ \hspace{11.8mm}  & 2.490 & 1.82 & 10.00 & 1.00 & 10.00 & 2.490 & 1.82 & 10.00 & 1.00 & 10.00 \\
Mistral $_{\mathrlap{\text{1}}}$ \hspace{11.8mm} & 2.470 & 3.74 & 11.00 & 21.01 & 11.00 & 2.470 & 3.74 & 11.00 & 21.01 & 11.00 \\
Mistral $_{\mathrlap{\text{3}}}$ \hspace{11.8mm}  & 1.490 & 19.66 & 21.00 & 36.86 & 21.00 & 1.500 & 18.35 & 20.00 & \textbf{36.18} & 20.00 \\
Mistral $_{\mathrlap{\text{5}}}$ \hspace{11.8mm} & 2.313 & 6.28 & 14.14 & 5.30 & 14.14 & 2.313 & 6.28 & 14.14 & 5.30 & 14.14 \\
Ensemble $_{\mathrlap{\text{0}}}$ \hspace{8.3mm}  & 2.310 & 6.47 & 12.13 & 4.36 & 12.13 & 2.050 & 9.83 & 18.00 & 7.90 & 18.00 \\
Ensemble $_{\mathrlap{\text{1}}}$ \hspace{8.3mm}  & 2.380 & 4.56 & 12.12 & 4.36 & 12.12 & 2.470 & 3.74 & 11.00 & 21.01 & 11.00 \\
Ensemble $_{\mathrlap{\text{3}}}$ \hspace{8.3mm}  & 1.480 & 20.94 & 23.00 & 27.97 & 23.00 & \textbf{1.440} & 21.86 & 25.00 & 26.61 & 25.00 \\
Ensemble $_{\mathrlap{\text{5}}}$ \hspace{8.3mm} & 2.484 & 1.87 & 10.10 & 1.03 & 10.10 & 2.505 & 1.85 & 10.10 & 1.02 & 10.10 \\

\midrule
 & & & & & \textbf{MBFC-2025} & & & & & \\
\midrule

 & & & & & \textbf{PLMs} & & & & & \\
\midrule
SVM$_{\mathrlap{\text{TF-IDF}}}$\hspace{16mm} & 0.448 & 25.88 & 62.69 & 27.18 & 28.55 & 0.443 & 26.20 & 63.18 & 27.14 & 28.76 \\
BERT$_{\mathrlap{\text{Base}}}$\hspace{14.8mm} & \textbf{0.303} & 33.07 & 72.14 & 38.94 & 33.82 & 0.328 & 32.32 & 70.15 & 48.02 & 32.92 \\
RoBERTa$_{\mathrlap{\text{Base}}}$\hspace{9.6mm} & 0.313 & \textbf{41.58} & 71.64 & 68.58 & 38.63 & 0.323 & 30.12 & 71.14 & 28.47 & 32.22 \\
DistilBERT$_{\mathrlap{\text{Base}}}$\hspace{7.8mm} & 0.308 & 33.31 & \textbf{72.64} & \textbf{49.01} & \textbf{34.08} & 0.318 & 32.92 & 71.64 & 48.59 & 33.62 \\
ALBERT$_{\mathrlap{\text{Base}}}$\hspace{11mm} & 0.343 & 31.93 & 69.65 & 37.98 & 32.75 & 0.333 & 32.19 & 70.15 & 38.13 & 32.95 \\
Ensemble\hspace{10.3mm} & 0.313 & 33.07 & 72.14 & 48.84 & 33.88 & 0.328 & 32.28 & 70.15 & 47.98 & 32.95 \\

\midrule
 & & & & & \textbf{LLMs} & & & & & \\
\midrule

LLaMA2 $_{\mathrlap{\text{0}}}$ $\hspace{9mm}$ & 1.953 & 10.65 & 19.67 & 7.99 & 19.67 & \textbf{1.947} & \textbf{11.25} & \textbf{20.67} & 8.46 & \textbf{20.67} \\
LLaMA2 $_{\mathrlap{\text{1}}}$ $\hspace{9mm}$ & 2.400 & 3.10 & 12.00 & 8.15 & 12.00 & 2.413 & 2.45 & 11.67 & 1.37 & 11.67 \\
LLaMA2 $_{\mathrlap{\text{3}}}$ $\hspace{9mm}$ & 2.413 & 2.45 & 11.67 & 1.37 & 11.67 & 2.413 & 2.45 & 11.67 & 1.37 & 11.67 \\
LLaMA2 $_{\mathrlap{\text{5}}}$ $\hspace{9mm}$ & 2.426 & 2.47 & 11.75 & 1.38 & 11.75 & 2.426 & 2.47 & 11.75 & 1.38 & 11.75 \\
Mistral $_{\mathrlap{\text{0}}}$ \hspace{11.8mm}  & 2.393 & 2.47 & 11.67 & 1.38 & 11.67 & 2.400 & 2.46 & 11.67 & 1.37 & 11.67 \\
Mistral $_{\mathrlap{\text{1}}}$ \hspace{11.8mm} & 2.400 & 4.37 & 12.67 & 21.71 & 12.67 & 2.400 & 4.37 & 12.67 & \textbf{21.71} & 12.67 \\
Mistral $_{\mathrlap{\text{3}}}$ \hspace{11.8mm}  & 2.420 & 2.44 & 11.67 & 1.36 & 11.67 & 2.420 & 2.44 & 11.67 & 1.36 & 11.67 \\
Mistral $_{\mathrlap{\text{5}}}$ \hspace{11.8mm} & 2.332 & 6.23 & 13.76 & 10.22 & 13.76 & 2.332 & 6.23 & 13.76 & 10.22 & 13.76 \\
Ensemble $_{\mathrlap{\text{0}}}$ \hspace{8.3mm}  & 2.193 & 6.66 & 14.67 & 6.38 & 14.67 & 2.053 & 10.26 & 18.33 & 8.51 & 18.33 \\
Ensemble $_{\mathrlap{\text{1}}}$ \hspace{8.3mm}  & 2.400 & 3.73 & 12.33 & 14.93 & 12.33 & 2.413 & 3.10 & 12.00 & 8.15 & 12.00 \\
Ensemble $_{\mathrlap{\text{3}}}$ \hspace{8.3mm}  & 2.413 & 2.44 & 11.67 & 1.37 & 11.67 & 2.420 & 2.44 & 11.67 & 1.36 & 11.67 \\
Ensemble $_{\mathrlap{\text{5}}}$ \hspace{8.3mm} & 2.352 & 4.25 & 12.75 & 6.92 & 12.75 & 2.426 & 2.47 & 11.75 & 1.38 & 11.75 \\

\bottomrule
\end{tabular}
}
\vspace{-0.5em}
\caption{Evaluation results for \emph{factuality} using hard and soft voting for each framework and in the ensemble.}
\label{Task 3B}
\end{table*}

\appendix

\end{document}